\documentclass[lettersize,journal]{IEEEtran}
\usepackage{amsmath,amsfonts}
\usepackage{algorithmic}
\usepackage{algorithm}
\usepackage{array}
\usepackage[caption=false,font=normalsize,labelfont=sf,textfont=sf]{subfig}
\usepackage{textcomp}
\usepackage{stfloats}
\usepackage{url}
\usepackage{verbatim}
\usepackage{graphicx}
\usepackage{cite}
\usepackage{multirow}
\usepackage{booktabs}
\usepackage[pagebackref=true,breaklinks=true,letterpaper=true,colorlinks,bookmarks=false]{hyperref}
\ifCLASSOPTIONcompsoc
  \usepackage[nocompress]{cite}
\else
  \usepackage{cite}
\fi
\ifCLASSINFOpdf
\else
\fi
\begin{document}
%
\title{Tolerating Annotation Displacement in Dense Object Counting via Point Annotation Probability Map}
%
%
%
%

\author{Yuehai~Chen, ~\IEEEmembership{Member,~IEEE,}
         Jing Yang, ~\IEEEmembership{Member,~IEEE,}
         Badong Chen, ~\IEEEmembership{Senior Member,~IEEE,}
         Shaoyi~Du, ~\IEEEmembership{Member,~IEEE}
         and Gang Hua,~\IEEEmembership{Fellow ,~IEEE} 
\IEEEcompsocitemizethanks{\IEEEcompsocthanksitem 
Yuehai~Chen and Jing Yang are with School of Automation Science and Engineering, Xi’an Jiaotong University, Xi’an 710049, China.

E-mail: cyh0518@stu.xjtu.edu.cn, jasmine1976@xjtu.edu.cn
\IEEEcompsocthanksitem Badong Chen and Shaoyi~Du is with Institute of Articial Intelligence and Robotics, Xi’an Jiaotong University, Xi’an, Shanxi 710049, China

E-mail: chenbd@mail.xjtu.edu.cn, dushaoyi@gmail.com
\IEEEcompsocthanksitem Gang Hua is with Wormpex AI Research LLC, Bellevue, WA 98004 USA.

E-mail: ganghua@gmail.com
\IEEEcompsocthanksitem Corresponding author: Jing Yang and Shaoyi Du.
}}

%
%

\markboth{Journal of \LaTeX\ Class Files,~Vol.~14, No.~8, August~2015}%
{Shell \MakeLowercase{\textit{et al.}}: Bare Advanced Demo of IEEEtran.cls for IEEE Computer Society Journals}
%



\IEEEtitleabstractindextext{%
\begin{abstract}
Counting objects in crowded scenes remains a challenge to computer vision. The current deep learning based approach often formulate it as a Gaussian density regression problem. Such a brute-force regression, though effective, may not consider the annotation displacement properly which arises from the human annotation process and may lead to different distributions. We conjecture that it would be beneficial to consider the annotation displacement in the dense object counting task. To obtain strong robustness against annotation displacement, generalized Gaussian distribution (GGD) function with a tunable bandwidth and shape parameter is exploited to form the learning target point annotation probability map, PAPM. Specifically, we first present a hand-designed PAPM method (HD-PAPM), in which we design a function based on GGD to tolerate the annotation displacement. For end-to-end training, the hand-designed PAPM may not be optimal for the particular network and dataset. An adaptively learned PAPM method (AL-PAPM) is proposed. To improve the robustness to annotation displacement, we design an effective transport cost function based on GGD. The proposed PAPM is capable of integration with other methods. We also combine PAPM with P2PNet through modifying the matching cost matrix, forming P2P-PAPM. This could also improve the robustness to annotation displacement of P2PNet. Extensive experiments show the superiority of our proposed methods.
\end{abstract}

}

\maketitle

\IEEEdisplaynontitleabstractindextext

%
\IEEEpeerreviewmaketitle

\ifCLASSOPTIONcompsoc
\IEEEraisesectionheading{\section{Introduction}\label{sec:introduction}}
\else
\fi

\IEEEPARstart{T}{he} counting task, consisting of crowd counting, vehicle counting, and general object counting, entails the estimation of target numbers within static images or video. This task has garnered heightened attention due to its extensive applicability in areas such as crowd analytics, traffic control, and video surveillance \cite{TIP1,TIP2,TIP3,TIP4,TIP5, region-aware}. The outbreak of the COVID-19 pandemic has further propelled its significance. Counting task contains a multitude of intensive counting scenes. In such contexts, the use of point annotation proves to be less labor-intensive in comparison to bounding-box annotation. Consequently, point annotation has gained widespread adoption within supervised methods \cite{first-denisty}.

\begin{figure}[t]
\begin{center}
\includegraphics[width=1.0\linewidth]{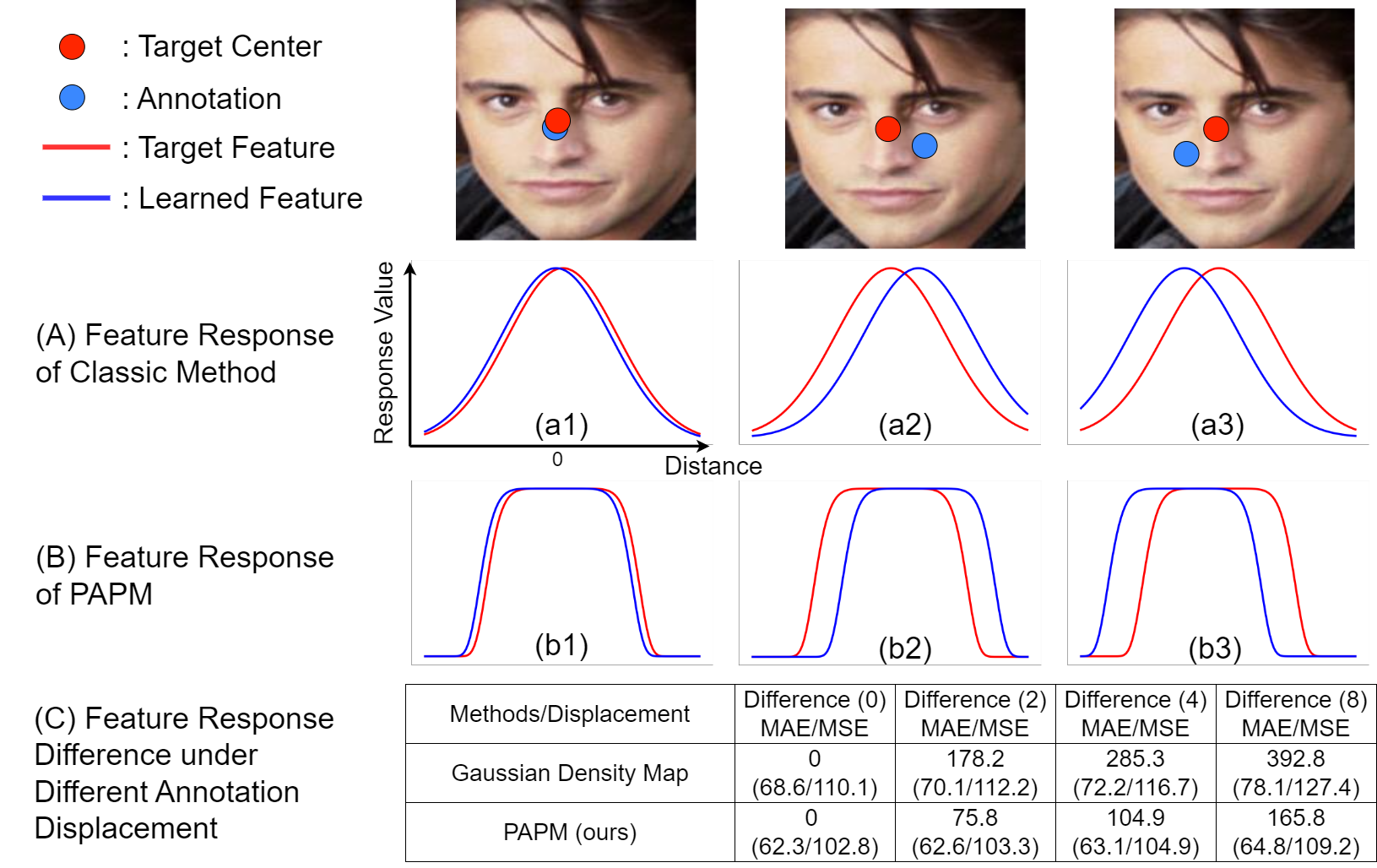}
\end{center}
   \caption{The qualitative results show the impact of annotation point offsets on (A) classic methods and (B) our proposed PAPM. (C) The quantitative results show the impact of annotation point offsets on classic method (Gaussian Density) and our proposed HD-PAPM. Specifically, we first generate noisy datasets by moving the annotation points by $\{2, 4, 8\}$ pixels in Part A \cite{MCNN} dataset. Then we train the vgg19 with different learning targets including Gaussian density map \cite{CSRNet} and HD-PAPM. For each image in the testing dataset, we calculate the pixel difference in density response maps by subtracting the results of model (offset 2, 4, or 8) from the results of model (offset 0). This difference reflects the impact of the annotation offset on the feature response.}
\label{fig:introduction1}
\end{figure}

The current counting methods which effectively utilize point annotation can be broadly classified into two categories: Gaussian density map supervision-based \cite{CSRNet, first-denisty} and point annotation supervision-based approaches \cite{DM-Count, UOT, p2pnet}. The former posits that the learned features conform to a Gaussian distribution, centered around the annotation points. The latter \cite{DM-Count, UOT} employs Euclidean distance as the transport cost function, with the underlying assumption that the closer a pixel is to its corresponding annotation point, the easier it is to transmit. These classic methods have made significant progress in counting tasks, but they still struggle with annotation displacement where they assume the features of pixels in proximity to the annotation point are of greater significance. We draw a schematic diagram in Figure \ref{fig:introduction1} to present the impact of annotation point offsets on these methods.

These offset annotations compel the network to learn features tied to the corresponding annotations, consequently impeding the acquisition of consistent features pertaining to the target. The red point in Figure \ref{fig:introduction1} (A) represents target center and its corresponding red curve represents the target feature that the network is supposed to learn. While the blue curve represents the feature of the annotation point area, which is the actual learned feature of the network. Upon the occurrence of an annotation shift, it can be observed in Figure \ref{fig:introduction1} (a2) and (a3) that there is almost no equal region between the target feature and the learned feature. This means that a small annotation displacement can have a significant impact on the consistency of the target feature and learned (annotation) feature. The quantitative results in Figure \ref{fig:introduction1} (C) also verify small annotation displacements bring big feature response difference for Gaussian density map method. The presence of inconsistency in feature space increases the difficulty for the model to learn representative features, resulting in a reduction in counting accuracy. Some recent methods have aimed to tackle annotation displacement by modeling \cite{NoiseCC} or correcting it iteratively \cite{ADSCNet}. However, these techniques are often tailored to specific architectures and may not be readily transferable to other approaches for improving their capacity to handle annotation displacement.

\begin{figure}[t]
\begin{center}
\includegraphics[width=1.0\linewidth]{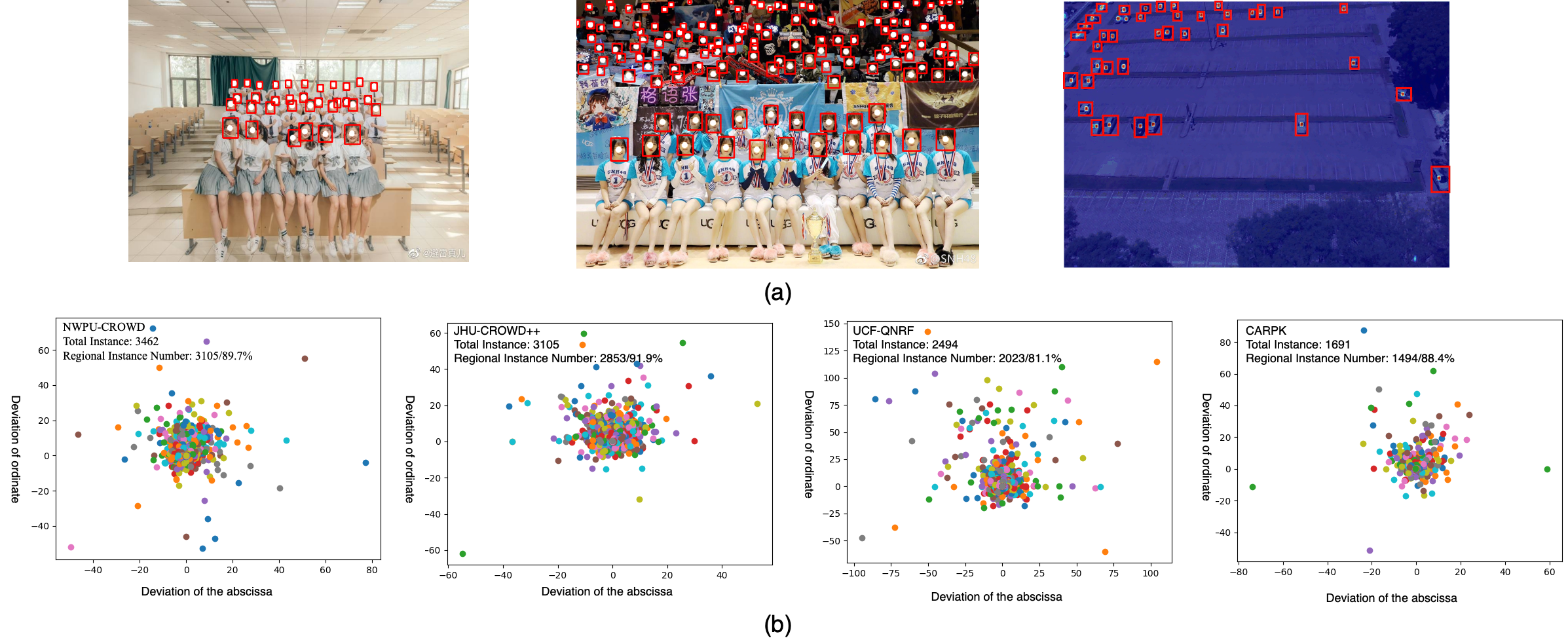}
\end{center}
   \caption{(a) Human may tend to annotate point at the center region of annotation box. (b) Most annotation point would be marked at the center region of annotation box (target region). We conducted an extensive annotation exercise utilizing diverse datasets \cite{NWPU, JHU2, UCF-QNRF, CARPK}, where we labeled annotations at the box level for numerous targets. Subsequently, we computed the distances between the annotation points and the centers of their corresponding annotation boxes.}
\label{fig:introduction2}
\end{figure}

Different to counting models, human have a strong tolerance for annotation displacement in counting task. Our observations indicate that when individuals are tasked with annotating a target, as illustrated in Figure \ref{fig:introduction2} (a), they tend to place their annotations at the central region of the annotation box. This tendency likely arises from the annotators' inclination to identify a potential target region, within which they strive to position their annotations \cite{elazary2008interesting}. Figure \ref{fig:introduction2} (b) presents an annotation analysis that a substantial concentration of annotation points is centered within the annotation boxes' central region, which corresponds to the target region. Given the subjective nature of human annotation, it's reasonable to anticipate some variability in the positioning of annotation points within the target region \cite{ADSCNet}. In essence, minor displacements of annotation points within the confines of the target region have a negligible impact on the counting task \cite{falk2019u}.

Motivated by this phenomenon, we propose a novel learning target, the point annotation probability map (PAPM), to enhance the model's resilience to annotation shifts. The central principle of PAPM is rooted in the assumption that each annotation point within the target region exerts an equal influence on the counting task. Specifically, PAPM assumes that the probability of people annotating in the target region is consistent. As a result, the feature response of PAPM should be equal across the target regions. As illustrated in the figure \ref{fig:introduction1} (b2) and (b3), when faced with the same degree of annotation shift, PAPM exhibits more equal regions between the target feature and its corresponding learned feature compared to classic methods. Therefore, as shown in Figure \ref{fig:introduction1} (C), PAPM can effectively reduce feature response inconsistencies caused by annotation shifts. This improves the model's robustness to annotation shifts. Our proposed PAPM is a general concept that can be easily incorporated with other methods, such as Gaussian density map \cite{CSRNet}, DM-Count \cite{DM-Count}, and P2PNet \cite{p2pnet}. Specifically, modifying the density map generation of Gaussian density map, the transmission cost function of DM-Count, and the matching cost matrix of P2PNet can improve robustness to annotation displacement, resulting in higher counting accuracy.

In summary, the contributions of the paper are three-fold:

\begin{itemize}

\item To address the challenge of classic counting methods struggling to adjust to annotation offsets, a novel learning target called the PAPM is introduced. The PAPM assumes that annotation points within the target region exert uniform influence on the counting task, accommodating the offsets of annotation points within this region.

\item The PAPM is a general concept that can be integrated with various methodologies. By combining PAPM with Gaussian density, DM-Count, and P2PNet, the resulting methods showcase marked enhancements in counting accuracy and resilience to annotation offsets when compared to their original counterparts.





\item The proposed approach demonstrates remarkable counting performances on ten diverse datasets covering three applications: crowd counting, vehicle counting, and general object counting.
\end{itemize}

\section{Related works}

\subsection{Crowd Counting Methods}

\noindent \textbf{Density map based crowd counting.} 
Lemptisky first uses Gaussian kernel to generate kernel density map from annotation dot maps as learning target \cite{first-denisty}. The density map alleviates the discrete nature of observation images (pixel grid) and points annotation (sparse dots). To generate a better density map learning target, some researchers adopt an adaptive kernel according to crowdedness or scene perspective to improve the quality of the learning target \cite{CSRNet,Adaptive-Density-Map1, Adaptive-Density-Map2}. ADMG design a learnable generation network to fuse density map of different variances as learning target \cite{ADMG}. Then, different network structures are proposed to deal with challenges in crowd counting, such as scale variation. 

From the standpoint that different kernels have receptive fields with different sizes, some researchers propose a multi-column convolution neural network to extract multi-scale features \cite{MCNN, switch-CNN}. 
Consider simplifying network architecture, some methods deploy single and deeper CNNs and consider combining features from different layers \cite{SaCNN, CSRNet}. SaCNN is a scale-adaptive CNN that combines feature maps extracted from multiple layers to perform the final density prediction \cite{SaCNN}. 
The attention-guided collaborative counting module proposed by AGCCM \cite{AGCCM} promotes collaboration between branches and has been shown to outperform state-of-the-art crowd counting methods.
Annotators typically position annotations within target regions, where slight displacements of annotated locations should be acceptable. However, numerous density map-based methods utilize Gaussian kernels to generate learning targets, which often lack consideration of annotation displacement. 

\noindent \textbf{Dot map based crowd counting.} 
Density maps are essentially intermediate representations that are constructed from an annotation dot map, whose optimal choice of bandwidth varies with the dataset and network architecture \cite{NoiseCC}. Thus, some point annotation directly based framework methods are proposed in crowd counting \cite{BL, DM-Count, UOT, GL}. The Bayesian loss (BL) uses a point-wise loss function between the ground-truth point annotations and the aggregated dot prediction generated from the predicted density map \cite{BL}. DM-Count considers density maps and dot maps as probability distributions and uses balanced OT to match the shape of the two distributions \cite{DM-Count}. GL \cite{GL} and UOT \cite{UOT} adopt unbalanced OT to improve the performance of DM-Count \cite{DM-Count}. These methods have demonstrated impressive performance. However, they encounter difficulties in effectively handling the displacement of annotation points. 

Existing methods for handling annotation offset in crowd counting, such as NoiseCC \cite{NoiseCC} and ADSCNet \cite{ADSCNet}, take different approaches to the problem. NoiseCC models annotation displacement as a random variable with a Gaussian distribution, and calculates the probability density function of crowd density values at each spatial location in the image. ADSCNet iteratively corrects annotations to account for labeling deviations. These methods are not general methods and cannot be easily integrated into other approaches to improve robustness to annotation displacement. In contrast, our proposed PAPM assumes that each annotation point within the target region exerts an equal influence on the counting task. Thus, the displacement of annotated locations in the target region is tolerable. The introduced PAPM functions as a common concept, is simpler in integration with a variety of methodologies than existing noise approaches. By combining PAPM with Gaussian density, DM-Count, and P2PNet, the resulting methods showcase marked enhancements in counting accuracy and resilience to annotation offsets when compared to their original counterparts.



\subsection{Generalize Gaussian Distribution}

The multivariate generalized Gaussian distribution (MGGD) has been extensively utilized in robust signal processing to address the challenges posed by severe noise changes and outliers \cite{GGD1, Correntropy-1}. The probability density function of the MGGD is expressed as \cite{MGGD}:

\begin{equation}
\begin{aligned}
k(\mathbf{x} ; \boldsymbol{\Sigma}, s, \sigma)=\frac{\Gamma\left(\frac{D}{2}\right)}{\pi^{\frac{D}{2}} \Gamma\left(\frac{D}{2 s}\right) 2^{\frac{D}{2 s}}} \frac{s}{\sigma^{D}|\boldsymbol{\Sigma}|^{\frac{1}{2}}} \\
\times \exp \left[-\frac{1}{2 \sigma^{2s}}\left(\mathbf{x}^{\top} \boldsymbol{\Sigma}^{-1} \mathbf{x}\right)^s\right],
\end{aligned}
\label{MGGD}
\end{equation}

\noindent where $D$ denotes the dimension of the probability space, where $\mathbf{x} \in \mathbb{R}^D$ represents a random vector. $\sigma$ is the bandwidth, $s>0$ is the shape parameter that controls the peakedness and the spread of the distribution, and $\boldsymbol{\Sigma}$ is a $D \times D$ symmetric positive scatter matrix. $\Gamma(D / 2s)=\int_0^{\infty} \mathrm{t}^{D/2s - 1} \mathrm{e}^{-\mathrm{t}} \mathrm{dt}$ denotes the Gamma function. The MGGD reduces to the multivariate Gaussian distribution when $s=1$, and $\boldsymbol{\Sigma}$ represents the covariance matrix. Additionally, when $s<1$, the distribution of the marginals becomes more peaky with heavier tails, whereas $s>1$ leads to a less peaky distribution with lighter tails \cite{MGGD}.

Prior research has demonstrated that the GGD can adapt to changes in sharpness near the origin through the use of a flexible parameter $s$, without altering the bandwidth $\sigma$ \cite{GGD1}. This property makes it well-suited for handling diverse types of noise. Therefore, by designing an appropriate GGD, it is possible to mitigate the effects of annotation displacement, leading to improved robustness. 


\section{Proposed Method}

\subsection{Overview}

\begin{figure*}
\begin{center}
\includegraphics[width=1.0\linewidth]{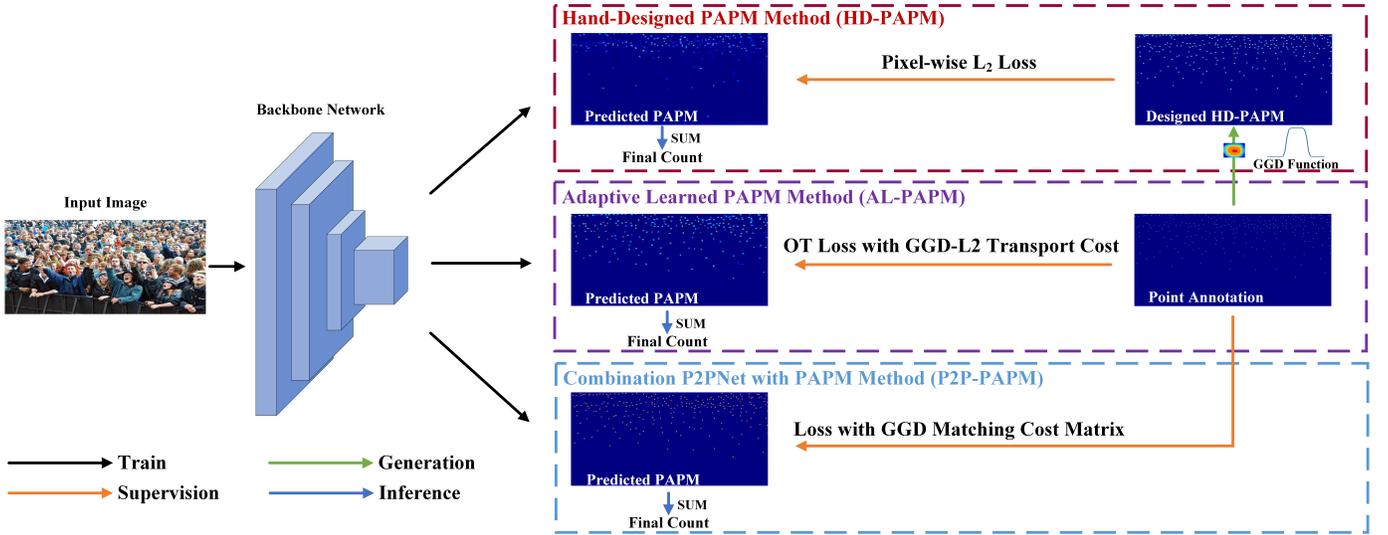}
\end{center}
   \caption{The overall framework of the proposed methods for object counting. Specifically, we combine the proposed PAPM with Gaussian density map method, DM-Count \cite{DM-Count}, and P2PNet \cite{p2pnet}, obtaining HD-PAPM, AL-PAPM, and P2P-PAPM methods.}
\label{fig:framework}
\end{figure*}

As discussed above, a substantial concentration of annotation points is centered within the target region of object. Given the subjective nature of human annotation, it's reasonable to anticipate some variability in the positioning of annotation points within the target region \cite{ADSCNet}. However, the classic methods \cite{CSRNet, DM-Count} is sensitive to annotation displacement, resulting in counting accuracy decline. To solve this problem, a novel learning target called PAPM is introduced. The core principle of PAPM is rooted in the assumption that each annotation point within the target region exerts an equal influence on the counting task. Specifically, PAPM assumes that the probability of people annotating in the target region is consistent, mitigating the impact of annotation displacement. Our proposed PAPM is a general concept that can be easily incorporated with other methods, such as Gaussian density map \cite{CSRNet}, DM-Count \cite{DM-Count}, and P2PNet \cite{p2pnet}.

In this work, we first combine PAPM with Gaussian density map, obtaining the hand-designed method (HD-PAPM). In the HD-PAPM method, we adopt a well-designed GGD kernel function to generate the PAPM as a learning target. Considering that hand-craft designed PAPM may not be an optimal learning target in deep learning, we combine PAPM with DM-Count \cite{DM-Count}. In this combining, we design an optimal transport framework to adaptively learn a better PAPM representation from point annotation in an end-to-end manner (AL-PAPM). Specifically, in the AL-PAPM method, we design a transport cost based on the GGD kernel function to tolerate the annotation displacement. Furthermore, the proposed PAPM can also be effectively combined with the P2PNet. Through the adaptation of the proposal matching cost matrix in the P2PNet method, the composite approach ``P2P-PAPM" effectively elevates the counting performance of the P2PNet method. The overall framework is presented in Figure \ref{fig:framework}.

\subsection{Hand-Designed Point Annotation Probability Map (HD-PAPM)}

As discussed above, the displacement of annotated locations in the target region should be tolerated. However, the common Gaussian function is too sharp at the origin, causing small annotation offsets to seriously affect the consistency between the target features and the actual learned features. To overcome this, we introduce the concept of GGD, which has been successful in dealing with noise \cite{GGD1, GGD2016}, into object counting.
Formally, given a set of $N$ input images $I_{1}, I_{2}$ $, \cdots, I_{N}$, we assume that each input image $I$ is associated with a set of $2 \mathrm{D}$ annotation points $P=\left\{p_{1}, \cdots, p_{n}\right\}$, where $p_{j}=(z_{j},y_{j})$ represents the position of $j$-$th$ annotated target, $n$ is the count number in input image $I$. Notably, input image $I$ is a dense real-value matrix, while the points annotation map is a sparse binary matrix (annotation points take the value 1 and 0 for otherwise). From an end-to-end training perspective, it is hard to directly adopt sparse point annotation $P$ as a learning target with per-pixel loss. To address this issue, as shown in the part framed by red dashed line in Figure \ref{fig:framework}, 
we design a GGD kernel function to convert point-level annotation $P=\left\{p_{1}, \cdots, p_{n}\right\}$ to head-level PAPM $A^{g t}$:

\begin{equation}
A\left(a\right)^{g t}=\sum_{j=1}^{n} k_{\sigma, s}\left(a, p_{j}\right),
\end{equation}

\noindent where $A^{g t}$ is the generated learning target PAPM, specifically, $a$ is the spatial location in the image, and $A\left(a\right)^{g t}$ is the corresponding value. $k_{\sigma, s}\left(a, p_{j}\right)=K \times \exp \left(- \left(\|a-p_{j}\|^{2} / 2 \sigma^{2}\right)^{s/2} \right)$ denotes a designed 2D distribution at the annotation $p_{j}$ of $j$-$th$ target, $K=\frac{1}{\pi \Gamma\left(\frac{1}{s}\right) 2^{\frac{1}{s}}} \frac{s}{\sigma^2|\boldsymbol{\Sigma}|^{\frac{1}{2}}} = \frac{s 2^s}{\pi \sigma^2 |\boldsymbol{\Sigma}|^{\frac{1}{2}} \Gamma\left(1/s \right)} $  is the normalized factor making $\sum_{\forall a} k_{\sigma, s} = 1$, specifically, $\Gamma(1 / s)=\int_0^{\infty} \mathrm{t}^{1/s-1} \mathrm{e}^{-\mathrm{t}} \mathrm{dt}$ is the Gamma function. $\boldsymbol{\Sigma}$ is a $2 \times 2$ symmetric positive scatter matrix. The bandwidth $\sigma$ and the shape parameter $s$ joint control the target regions where point annotation is likely to be marked. 


\begin{figure}[htbp]
    \begin{center}
    \includegraphics[width=1.0\linewidth]{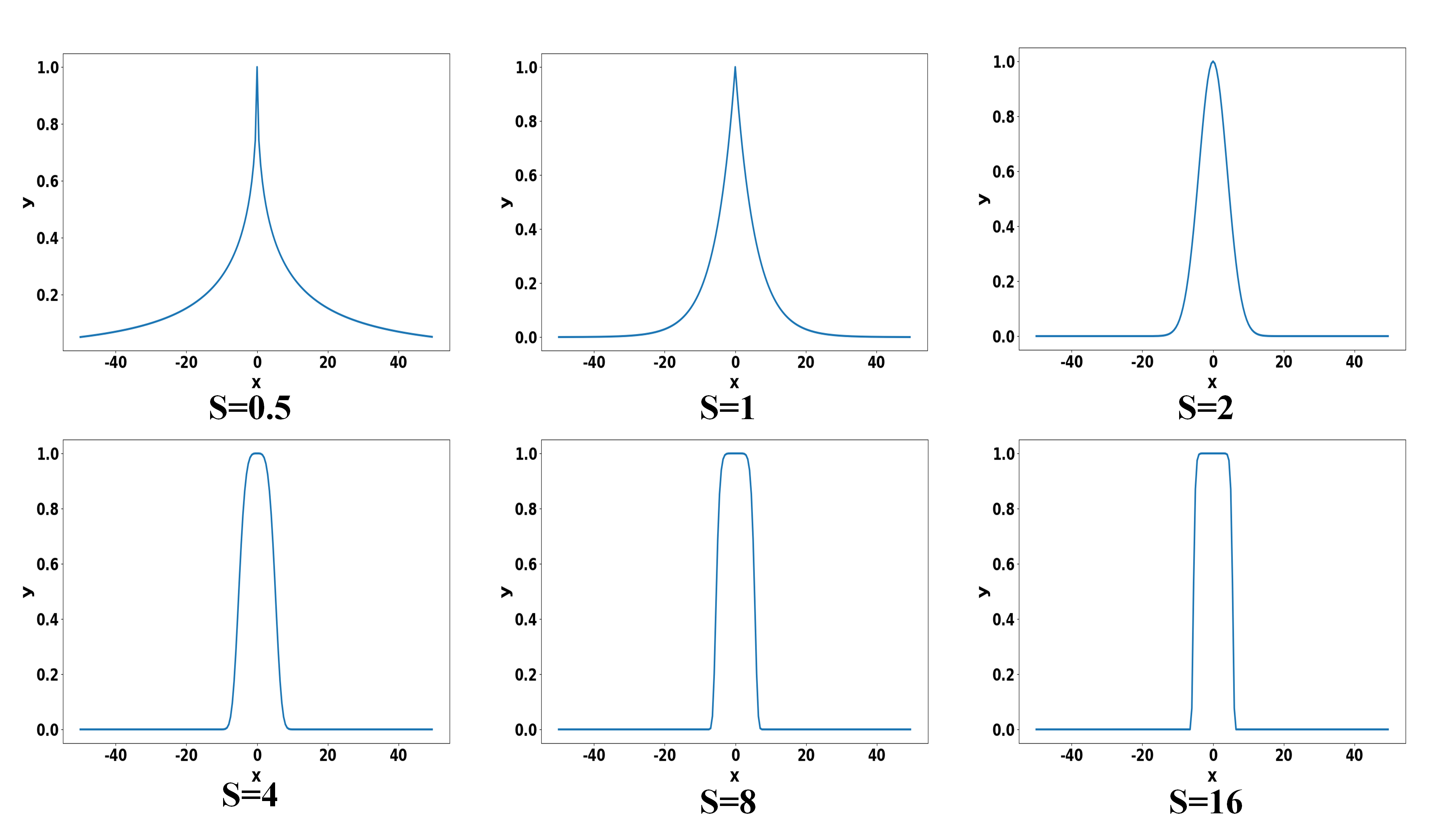}
    \end{center}
   \caption{Visualization of GGD in HD-PAPM with different parameter $s$. The x-axis in represents the distance between two pixels, and the y-axis represents the response value.}
    \label{fig:s}
    \end{figure}

To better understand, Figure \ref{fig:s} shows the plots of the designed GGD kernel functions with different shape parameters $s$. It could be observed that the GGD kernel function could smooth the origin surface by changing the shape parameter $s$. The larger $s$ is, the smoother the surface near the origin is. This means that GGD with large shape parameter $s$ treats the pixels in the center region similar. As a result, the GGD kernel function can tolerate the annotation displacement. When $s = 2$, GGD function converts to the Gaussian function.

In the HD-PAPM method, we adopt the per-pixel $L_2$ loss to optimize the network model:

\begin{equation}
L_{2}=\frac{1}{2 N} \sum_{i=1}^{N} \left\|A_{i}^{g t}\left(a\right)-A_{i}^{e s t}\left(a\right)\right\|^{2}_{2},
\end{equation}

\noindent where $A_{i}^{est}(a)$ is the estimated PAPM of training image $I_{i}$, which is generated from neural network model. $A_{i}^{gt}(a)$ is the target PAPM, $N$ is the number of training image and $\|\|^{2}_{2}$ is $L_2$ loss function.

\subsection{Adaptive Learned Point Annotation Probability Map via Optimal Transport (AL-PAPM)}
From the standpoint of end-to-end training, the hand-designed PAPM may not be optimal for the particular network architecture and particular dataset. Thus, we consider how to adaptively learn a better PAPM representation. In the counting task, we assume that the annotation process on each target obeys a potential distribution, and consider the ground-truth point annotation to be an observation of the potential distribution. To seek the potential distribution, as shown in the part framed by the purple dashed line in Figure \ref{fig:framework}, 
we naturally consider minimizing the distance between predicted PAPM and ground-truth point annotations through optimal transport (OT) \cite{OT-original}. Specifically, we build upon the optimal transport framework proposed in DM-Count \cite{DM-Count} and modify the transport cost function to suit the counting task.

\begin{figure}[t]
    \begin{center}
    \includegraphics[width=1.0\linewidth]{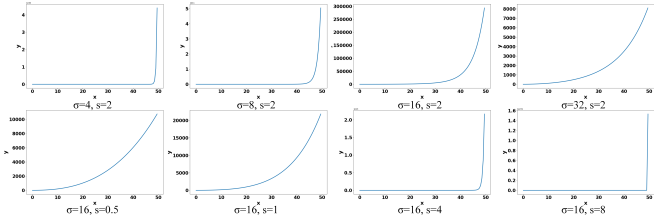}
    \end{center}
   \caption{Visualization of the transport cost function in AL-PAPM with different parameters $\sigma$ and $s$. The x-axis represents the distance between two pixels, and the y-axis represents the transmission cost.}
    \label{fig:als}
    \end{figure}

As discussed above, minor displacements of annotation points within the target region have a negligible impact on the counting task \cite{falk2019u}. The annotations in the target regions of object can be equally effective to the counting task. In conclusion, the effectiveness scope of annotation is local rather than global. GGD function is proved to meet the locality requirement \cite{Correntropy-1}. 
Figure \ref{fig:als} illustrates that the bandwidth parameter controls the range of the target regions: a larger bandwidth results in smaller target regions. Similarly, the shape parameter controls the shape of the transport cost, with larger shape parameters resulting in larger target regions. Inspired by the local metric property of the GGD function, we extend the GGD function to OT to improve the robustness to annotation displacement through setting suitable bandwidth $\sigma$ and shape parameter $s$.

Let $P=\left\{p_{i}\right\}_{i=1}^{n}$ be the ground-truth point annotation ($p_{i}$ is the annotation position, $n$ is the number of annotation) and $A=\left\{a_{i}\right\}_{i=1}^{m}$ be the PAPM ($a_{i}$ is the position of pixel , $m$ is the number of pixels), respectively. Note that OT distance requires that the total mass of the input measures should be equal, otherwise, there is no feasible solution \cite{OT-theory}. Thus, following DM-Count \cite{DM-Count}, we turn the two measures into probability distribution functions by dividing them by their respective total mass.
Specifically, we consider the ground-truth point annotation distribution $\frac{P}{\|P\|_{1}}$ to be separable, and divide them into different probability masses. Then these different probability masses would be transported to different locations to form the point annotation probability distribution map $\frac{A}{\|A\|_{1}}$, through minimizing the transport cost $\ell_{\mathbf{C}}$ :

\begin{equation}
    \ell_{\mathbf{C}}(\frac{P}{\|P\|_{1}}, \frac{A}{\|A\|_{1}}) \stackrel{\bigtriangleup}{=} \min _{\mathbf{T} \in \mathbf{U}(P, A)}\langle\mathbf{C}, \mathbf{T}\rangle \stackrel{\bigtriangleup}{=} \sum_{i, j} \mathbf{C}_{i, j} \mathbf{T}_{i, j},
    \end{equation}

\noindent where $\|\cdot\|_{1}$ denote the $L_{1}$ norm of a vector, $\mathbf{C} \in \mathbb{R}_{+}^{n \times m}$ is the transport cost matrix, whose item $C_{i j}=c(p_{i},a_{j})$ measures the cost for moving probability mass on pixel $p_{i}$ to pixel $a_{j}$. $\left\{\mathbf{T} \in \mathbb{R}_{+}^{n \times m}: \mathbf{T} \mathbf{1_{m}}=\mathbf{1_{n}}, \mathbf{T}^{T} \mathbf{1_{n}}=\mathbf{1_{m}}\right\}$ ($\mathbf{T}$ is the transport matrix, which assigns probability masses at each location $p_{i}$ to $a_{j}$ for measuring the cost. $\mathbf{U}$ is the set of all possible ways to transport probability masses from $P$ to $A$.

Note that annotators would tend to mark annotation in the target regions of object and a bit of labeling position deviation is reasonable. Therefore, the point annotations should be transported to the pixels in the target region rather than other regions. This transmission relationship is embodied in the OT problem as: the cost of transporting the probability mass of point annotation to pixels in the target regions should be low and high outside the target regions. To reflect this transmission relationship, we extend the GGD kernel function into OT. To ensure that the transmission cost be 0 when the distance between annotation $p_{i}$ to pixel $a_{j}$ is equal to 0, we combine the GGD function and Euclidean distance to obtain the GGD-L2 combination cost function $c(p_{i},a_{j})$:

\begin{equation}
\begin{aligned}
c(p_{i},a_{j}) 
=\frac{ \left\|p_{i}-a_{j}\right\|^{2} }{\kappa_{\sigma, s}(p_{i}, a_{j})} 
= \frac{\left\|p_{i}-a_{j}\right\|^{2}} {\exp \left(- \left(\|p_{i}-a_{j}\|^{2}/{2 \sigma^{2}}\right) ^{s/2}
\right)} &\\,
\end{aligned}
\end{equation}

\noindent where $\kappa_{\sigma, s}(p_{i}, a_{j})$ is a GGD kernel, whose variance $\sigma$ and parameter $s$ joint control the target regions.

\begin{figure}[ht]
\begin{center}
\includegraphics[width=1.0\linewidth]{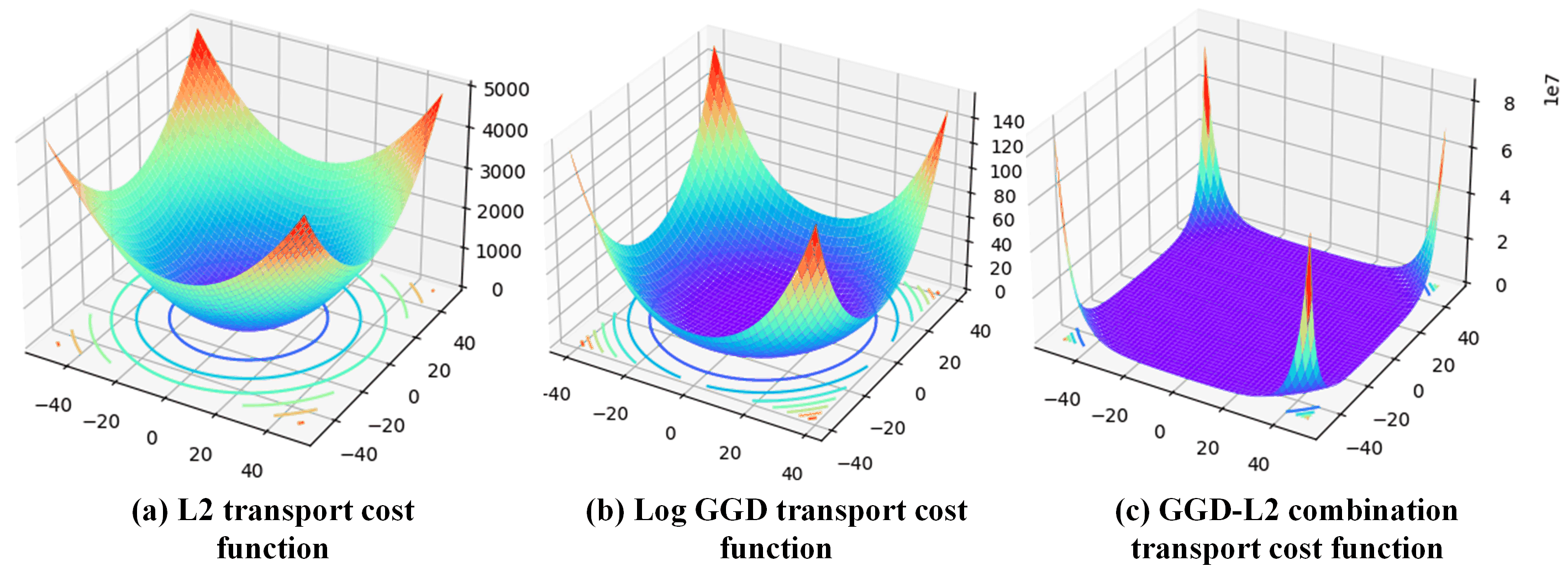}
\end{center}
   \caption{Comparison of transport cost functions based on Euclidean distance, log GGD kernel ($\sigma$=16, s=4), and the combination of the GGD kernel and L2-Square distance (GGD-L2 combination).}
\label{fig:kernel function}
\end{figure}

\textbf{Discussion.}
A typical transport cost function in OT is the Euclidean distance between two pixels $L^{i j}_{2}=\left\|p_{i}-a_{j}\right\|_{2}^{2}$. As shown in Figure \ref{fig:kernel function} (a), Euclidean distance $L^{i j}_{2}=\left\|p_{i}-a_{j}\right\|_{2}^{2}$ is very smooth but without boundary. It is a global metric that could not reflect the above-mentioned transmission relationship in the target regions. Compared with the Euclidean distance, the GGD-L2 combination cost function is local \cite{Correntropy-1}. As shown in Figure \ref{fig:kernel function} (c), the GGD-L2 combination transport cost function is bounded which could build completely different transport costs for inside and outside the target regions. Moreover, in Figure \ref{fig:kernel function} (c), we could observe that the transport cost is all low in the target region. This indicates that our designed transport cost function treats the pixels in the target region equally. Thus, our method could tolerate the displacement of the annotated locations in the target region.

Considering that our ultimate goal is to get the final count of observation image, following DM-Count \cite{DM-Count}, we add similarity counting loss $\ell_{S}(P, A) = \left|\|P\|_{1}-\|A\|_{1}\right| + \|P\|_{1}\left\|\frac{P}{\|P\|_{1}}-\frac{{A}}{\|{A}\|_{1}}\right\|_{1}$ to make the predicted count $\|A\|_{1}$ close to the ground-truth count $\|P\|_{1}$. We finally combine the optimal transport loss and the similarity counting loss to obtain the overall loss:

\begin{equation}
\ell(P, A)=\lambda_{1} \ell_{\mathbf{C}}(\frac{P}{\|P\|_{1}}, \frac{A}{\|A\|_{1}})+ \ell_{S}(P, A),
\end{equation}

\noindent where $\lambda_{1}$ is a hyper-parameter for the OT loss. Note that, the idea of tolerating the annotation displacement is flexible. We believe this idea could be plugged into other methods, such as GL \cite{GL} and UOT \cite{UOT}, and would improve results. The related experiment results have been presented in Table \ref{tab:different backbones}.

\subsection{Combination P2PNet with Point Annotation Probability Map (P2P-PAPM)}

Our proposed PAPM is a general concept that can be easily incorporated with other methods for tolerating annotation displacement. For example, the proposed PAPM can be effectively combined with the P2PNet \cite{p2pnet}. As shown in the part framed by the blue dashed line in Figure \ref{fig:framework}, Through the adaptation of the proposal matching cost matrix in the P2PNet method, the composite approach "P2P-PAPM" effectively elevates the counting performance. 

The outputs of P2PNet are predicted point proposals $\hat{P}=\{\hat{p}_1, ..., \hat{p}_{n_1}\}$ and corresponding confidence scores $\hat{C}=\{\hat{c}_1, ..., \hat{c}_{n_1}\}$, where ${n_1}$ refers to the number of predicted point proposal. To train the P2PNet model, we need to match the predicted point proposals and ground truth $P=\{p_1, ..., p_n\}$ by one-to-one, and the unmatched predicted points are considered to the ``background” class. $n$ refers to the number of ground truth points, which is smaller than ${n_1}$ to ensure each ground truth matches a prediction point. Next, we need to find a bipartite matching between predictions and ground truth with the lowest cost. A straightforward way in P2PNet \cite{p2pnet} is to take the $L_2$ distance and confidence as matching cost matrix $\mathcal{D}_{L2}$:

\begin{equation}
\mathcal{D}_{L2}({P}, \hat{{P}})=\left(\tau\left\|p_i-\hat{p}_j\right\|^2_2-\hat{c}_j\right)_{i \in n, j \in {n_1}}
\end{equation}

\noindent where $\|\cdot\|^2_2$ denotes to the $L_2$ distance, and $\hat{c}_j$ is the confidence score of the proposal $\hat{p}_j$. $\tau$ is a weight term to balance the effect from the pixel distance \cite{p2pnet}. 

Based on the $L_2$ cost matrix $\mathcal{D}_{L2}$, P2PNet \cite{p2pnet} utilizes the Hungarian \cite{kuhn1955hungarian} to implement one-to-one matching. However, we find that merely taking the the $L_2$ cost matrix $\mathcal{D}_{L2}$ with confidence could not tolerate annotation displacement. Because $L_2$ is sensitive to distance, the offset of annotations will increase the matching cost, resulting in unsatisfactory matching results. However, minor displacements of annotation points within the target region have a negligible impact on the counting task \cite{falk2019u}. Therefore, we introduce GGD-based cost matrix $\mathcal{D}_{ggd}$: 

\begin{equation}
\begin{aligned}
&\\\mathcal{D}_{ggd}({P}, \hat{{P}})=\left(\tau
\frac{\left\|p_i-\hat{p}_j\right\|^2}{\kappa_{\sigma, s}(p_i, \hat{p}_j)}-\hat{c}_j\right)_{i \in n, j \in {n_1}} 
&\\ = \left(\tau\frac{\left\|p_{i}-\hat{p}_j\right\|^{2}} {\exp \left(- \left(\|p_{i}-\hat{p}_j\|^{2}/{2 \sigma^{2}}\right) ^{s/2}
\right)}-\hat{c}_j\right)_{i \in n, j \in {n_1}}
\end{aligned}
\end{equation}

\noindent where $\kappa_{\sigma, s}(p_{i}, a_{j})$ is a GGD kernel, whose variance $\sigma$ and parameter $s$ joint control the most possibly annotated regions. The hyper parameters $\sigma$ and $s$ setting in P2P-PAPM are the same to those in AL-PAPM. $\tau=1e-5$ is a weight term to balance the effect from the pixel distance. 

The proposed GGD-based cost matrix $\mathcal{D}_{ggd}$ is similar to GGD-L2 combination transpont cost function in Figure \ref{fig:als}. Joint controlling variance $\sigma$ and parameter $s$ can reduce the matching cost of the annotation point and its corresponding target area proposal point to be similar. The similar matching cost in the target region means that the model can tolerate the offset of annotation points in the target region.

From the perspective of the ground truth points, let us use a permutation $\xi$ of $\{1, \ldots, {n_1}\}$ to represent the optimal matching result, i.e., $\xi=\Omega({P}, \hat{P}, \mathcal{D}_{ggd})$, where $\Omega({P}, \hat{P}, \mathcal{D}_{ggd})$ is the one-to-one matching strategy. That is to say, the ground truth point $p_i$ is matched to the proposal $\hat{p}_{\xi(i)}$. Furthermore, those matched proposals (positives) could be represented as a set $\hat{{P}}_{\text {pos }}=\left\{\hat{p}_{\xi(i)} \mid i \in\{1, \ldots, n\}\right\}$, and those unmatched proposals in the set $\hat{{P}}_{n e g}=\left\{\hat{p}_{\xi(i)} \mid i \in\{n+1, \ldots, {n_1}\}\right\}$. Following \cite{p2pnet}, after the ground truth targets have been obtained, we calculate the distance loss $\ell_{dis}$ to supervise the point regression, and use Cross Entropy loss $\ell_{cls}$ to train the proposal classification. The final loss function $\ell_{p2p}$ is the summation of the above two losses, which is defined as:

\begin{equation}
    \ell_{c l s}=-\frac{1}{{n_1}}\left\{\sum_{i=1}^n \log \hat{c}_{\xi(i)}+\lambda_2 \sum_{i=n+1}^{n_1} \log \left(1-\hat{c}_{\xi(i)}\right)\right\} 
\end{equation}
\begin{equation}
    \ell_{dis}=\frac{1}{n} \sum_{i=1}^n\left\|p_i-\hat{p}_{\xi(i)}\right\|_2^2
\end{equation}
\begin{equation}
    \ell_{p2p}=\ell_{c l s}+\lambda_3 \ell_{dis}
\end{equation}

\noindent where $\lambda_2=0.5$ is a reweight factor for negative proposals, and $\lambda_3=2e-4$ is a weight term to balance the effect of the regression loss.

\section{Experiments}

In this section, we present experiments evaluating the proposed methods, HD-PAPM, AL-PAPM, and P2P-PAPM. We first present a detailed experimental setup including datasets, network architecture, and evaluate metrics. Then, we compare the proposed methods with recent state-of-the-art approaches. Finally, we conduct ablation studies to verify the effectiveness of the proposed PAPM.

\subsection{Experimental Setups}
\noindent \textbf{Applications \& Datasets.}
We conduct experiments on three applications: crowd counting, vehicle counting, and general object counting. 
For \textbf{crowd counting}, six datasets are used for evaluation, including ShanghaiTech (ShTech) A and B \cite{MCNN}, UCF\_CC\_50 \cite{UCF-CC-50}, UCF-QNRF \cite{UCF-QNRF}, JHU-CROWD++ \cite{JHU2} and NWPU-Crowd \cite{NWPU}. ShTech A consists of 482 images with crowd numbers varying from 33 to 3139, and ShTech B contains 716 images with fewer crowd numbers from 9 to 578. UCF\_CC\_50 is an extremely dense crowd dataset including 50 images with an average number of 1280. UCF-QNRF, JHU-CROWD++, and NWPU-Crowd are three large-scale datasets that contain 1535, 4250, and, 5109 high-resolution images with very large crowds. Note that the ground truth for test images set in NWPU-Crowd is not released and researchers could only submit their results online for evaluation.
For \textbf{vehicle counting}, TRANCOS \cite{V3}, PUCPR+ \cite{CARPK} and CARPK \cite{CARPK} are used for evaluation. TRANCOS contains 1244 images in traffic with vehicle numbers varying from 9 to 107. PUCPR+ and CARPK are used to count parking cars. PUCPR+ contains only 125 images with vehicle numbers from 0 to 331, while CARPK is a large dataset with 1448 images.
The proposed methods are also evaluated on \textbf{general object counting} task on DOTA \cite{DOTA}, which contains more than one semantic class. For DOTA, following the ADMG work \cite{Adaptive-Density-Map2}, we first use 690 images with 6 classes (Large-vehicle, Helicopter, Plane, Ship, Small-vehicle, and Storage tank) with object numbers larger than 10, denoted as ``6 classes''. We also use 1869 high-resolution images with 18 classes with number varying from 0 to 1940, denoted as ``18 classes''.

\noindent \textbf{The Network Architecture.}
Our methods are denoted as hand-craft designed Point Annotation Probability Map (HD-PAPM, Section 3 B), adaptive learned Point Annotation Probability Map via designed OT loss function (AL-PAPM, Section 3 C) and combination P2PNet with Point Annotation Probability Map (P2P-PAPM, Section 3 D). 
To demonstrate the effectiveness of the proposed methods, we follow BL \cite{BL} and adopt VGG19 as the backbone network. Separate HD-PAPM and AL-PAPM means the network backbone is VGG19. To verify the generality of our method, we also embed our HD-PAPM, AL-PAPM into CSRNet \cite{CSRNet}, M-SFANet \cite{M-SFANet} and MAN \cite{MAN} network architectures. The bandwidths of Gaussian density map in object counting are set the same to \cite{Adaptive-Density-Map2}. 

    \begin{table*}
\caption{Results on the ShanghaiTech, UCF\_CC\_50, UCF-QNRF, NWPU and JHU-CROWD++ datasets. }
\begin{center}
\begin{tabular}{l|ll|ll|ll|ll|ll|ll}
 \hline 
 \multicolumn{1}{l|} {Method} & \multicolumn{2}{l|}{ ShTech A } & \multicolumn{2}{l|}{ ShTech B } & \multicolumn{2}{l|}{ UCF\_CC\_50 } & \multicolumn{2}{l|}{ UCF-QNRF } & \multicolumn{2}{l|}{ NWPU} & \multicolumn{2}{l}{ JHU-CROWD++}\\
& {MAE} & MSE & MAE & MSE & MAE & MSE & MAE & MSE & MAE & MSE & MAE & MSE \\
\hline MCNN\cite{MCNN} & $110.2$ & $173.2$ & $26.4$ & $41.3$ & $377.6$ & $509.1$ & $277.0$ & $426.0$ & $232.5$ & $714.6$ & $188.9$ &$483.4$ \\
Switch-CNN\cite{switch-CNN} & $90.4$ & $135.0$ & $21.6$ & $33.4$ & $318.1$ & $439.2$ & 228.0 & 445.0 & $-$ & $-$ & $-$ & $-$\\
CAN\cite{CAN} & $62.3$ & 100.0 & $7.8$ & $12.2$ & $212.2$ & $243.7$ & 107.0 & 183.0 & $106.3$ & ${386.5}$ & $-$ & $-$\\
SFCN\cite{SFCN} & $67.0$ & $104.5$ & $8.4$ & $13.6$  &  $258.4$ & $334.9$ & $102.0$ & $171.4$ & $105.7$ & $424.1$ & $77.5$ & $297.6$\\
ADSCNet\cite{ADCrowdNet} & $55.4$ & $97.7$ & $\textbf{6.4}$ & $11.3$  & $198.4$ & $267.3$ & $\textbf{71.3}$ & $\textbf{132.5}$ & $-$ & $-$ & $-$ & $-$\\
SASNet\cite{SASNet} & $\textbf{53.6}$ & $\textbf{88.4}$ & $\textbf{6.4}$ & $\textbf{10.0}$  & $\textbf{161.4}$ & $\textbf{234.5}$ & $85.2$ & $147.3$ & $-$ & $-$ & $-$ & $-$ \\
BL\cite{BL} & $62.8$ & $101.8$ & $7.7$ & $12.7$ & $229.3$ & $308.2$ & $88.7$ & $154.8$  & $105.4$ & $454.0$ & $75.0$ & $299.9$\\
NoiseCC\cite{NoiseCC} & $61.9$ & $99.6$ & $7.4$ & $11.3$  & $-$ & $-$ & $85.8$ & $150.6$ & $96.9$ & $534.2$ & $67.7$ & $258.5$\\
GL\cite{GL} & $61.3$ & $95.4$ & $7.3$ & $11.7$  & $-$ & $-$ & $84.3$ & $147.5$ & $\textbf{79.3}$ & $\textbf{346.1}$ & $59.9$ & $259.5$\\
ChfL \cite{chfL} & $57.5$ & $94.3$ & $6.9$ & $11.0$  & $-$ & $-$ & $80.3$ & $137.6$ & $76.8$ & $343.0$ & $57.0$ & $325.7$\\
\hline vgg19+Gaussian Density & $68.6$ & $110.1$ & $8.5$ & $13.9$ & $251.6$ & $331.3$ & $106.8$ & $183.7$ & $135.1$ & $442.8$ & $75.4$ & $292.1$ \\
vgg19+HD-PAPM(ours) & $62.3$ & $101.2$ & $7.6$ & $12.3$  & $220.1$ & $300.1$ & $97.2$ & $161.4$ & $129.2$ & $402.2$ &$69.1$ & $270.2$\\
vgg19+DM-Count & $59.7$ & $95.7$ & $7.4$ & $11.8$  & $211.0$ & $291.5$ & $85.6$ & $148.3$ & $88.4$ & $357.6$ &$ 68.4$ & $283.3$\\
vgg19+AL-PAPM(ours)  & $\textbf{57.1}$ & $\textbf{92.5}$ & $\textbf{7.0}$ & $\textbf{10.9}$ & $\textbf{195.7}$ & $\textbf{273.5}$ & $\textbf{81.2}$ & $\textbf{141.9}$ & $\textbf{79.7}$ & $\textbf{347.8}$ & $\textbf{56.5}$ & $\textbf{251.5}$\\
\hline 
CSRNet\cite{CSRNet}+Gaussian Density & $68.2$ & $115.0$ & $10.6$ & $16.0$ & $266.1$ & $397.5$ & $120.3$ & $208.5$ & $121.3$ & $387.8$ & $85.9$ & $309.2$\\
CSRNet+HD-PAPM(ours)  & $63.2$ & $102.8$ & $8.8$ & $14.2$ & $238.6$ & $362.5$ & $108.7$ & $184.9$ & $111.9$ & $353.3$ & $76.8$ & $284.2$\\
CSRNet+DM-Count & $61.3$ & $99.7$ & $8.4$ & $13.5$ & $228.4$ & $340.1$ & $103.6$ & $180.6$ & $92.4$ & $377.5$ & $72.3$ & $294.0$\\
CSRNet+AL-PAPM(ours)  & $\textbf{58.1}$ & $\textbf{94.7}$ & $\textbf{7.8}$ & $\textbf{13.5}$ & $\textbf{202.7}$ & $\textbf{291.3}$ & $\textbf{95.6}$ & $\textbf{162.7}$ & $\textbf{84.5}$ & $\textbf{355.9}$ & $\textbf{62.7}$ & $\textbf{262.8}$\\
\hline 
M-SFANet\cite{M-SFANet}+Gaussian Density & $59.7$ & $95.7$ & $6.8$ & $11.9$ & $162.3$ & $276.8$ & $85.6$ & $151.2$ & $87.5$ & $395.6$ & $69.6$ & $277.6$\\
M-SFANet+HD-PAPM(ours)  & $58.6$ & $93.8$ & $6.7$ & $111.5$ & $158.4$ & $264.1$ & $83.8$ & $147.1$ & $82.9$ & $371.2$ & $62.5$ & $253.4$\\
M-SFANet+DM-Count & $57.8$ & $92.4$ & $7.6$ & $12.6$ & $160.2$ & $272.3$ & $82.2$ & $145.7$ & $81.8$ & $345.0$ & $62.8$ & $258.8$\\
M-SFANet+AL-PAPM(ours)  & $\textbf{55.2}$ & $\textbf{89.8}$ & $\textbf{6.7}$ & $\textbf{11.4}$ & $\textbf{156.2}$ & $\textbf{258.4}$ & $\textbf{80.4}$ & $\textbf{142.3}$ & $\textbf{76.2}$ & $\textbf{323.1}$ & $\textbf{55.2}$ & $\textbf{239.3}$\\
\hline
MAN*\cite{MAN} & $56.2$ & $89.9$ & $-$ & $-$  & $-$ & $-$ & $78.0$ & $138.0$ & $76.5$ & $323.0$ & $52.7$ & $223.2$\\
MAN+Gaussian Density & $59.8$ & $96.4$ & $-$ & $-$  & $-$ & $-$ & $85.7$ & $149.8$ & $86.2$ & $382.1$ & $62.8$ & $255.1$\\
MAN+HD-PAPM(ours)  & $57.2$ & $93.8$ & $-$ & $-$  & $-$ & $-$ & $84.1$ & $145.4$ & $83.6$ & $364.4$ & $59.9$ & $240.7$\\
MAN+DM-Count & $55.7$ & $91.7$ & $-$ & $-$  & $-$ & $-$ & $80.4$ & $141.2$ & $77.5$ & $333.7$ & $56.0$ & $230.2$\\
MAN+AL-PAPM(ours) & $53.2$ & $85.6$ & $7.1$ & $11.2$ & $-$ & $-$ & $\textbf{76.4}$ & $\textbf{136.1}$ & $\textbf{73.2}$ & $\textbf{315.2}$ & $\textbf{52.1}$ & $\textbf{214.4}$\\
\hline 
P2PNet \cite{p2pnet} & ${52.7}$ & ${85.1}$ & ${6.2}$ & ${9.9}$  & $-$ & $-$ & $85.3$ & $154.5$ & $77.4$ & $362.0$ & $-$ & $-$\\
P2P-PAPM(ours)  & $\textbf{51.2}$ & $\textbf{83.5}$ & $\textbf{6.0}$ & $\textbf{9.2}$  & $-$ & $-$ & $\textbf{82.8}$ & $\textbf{145.2}$ & $\textbf{74.8}$ & $\textbf{355.4}$ & $-$ & $-$ \\
\hline
\end{tabular}
\end{center}
MAN* means the reproduced results that we use the official codes provided by MAN\cite{MAN} paper to get.
\label{table:Comparisons}
\end{table*}

\noindent \textbf{Evaluation Metrics.}
The widely used mean absolute error (MAE) and the mean squared error (MSE) are adopted to evaluate the performance. The MAE and MSE are defined as follows:

\begin{equation}
M A E=\frac{1}{N} \sum_{i=1}^{N}\left|\hat{g}_{i}-g_{i}\right|,
\end{equation}

\begin{equation}
M S E=\sqrt{\frac{1}{N} \sum_{i=1}^{N}\left|\hat{g}_{i}-g_{i}\right|^{2}},
\end{equation}
where $N$ is the number of test images, $\hat{g}_{i}$ and $g_{i}$ are the estimated count and the ground-truth, respectively.

\subsection{Crowd Counting}
Our proposed PAPM is a general concept that can be easily incorporated with other methods, such as Gaussian density map \cite{CSRNet}, DM-Count \cite{DM-Count}, and P2PNet \cite{p2pnet}, forming HD-PAPM, AL-PAPM, and P2P-PAPM. The proposed HD-PAPM and AL-PAPM is general, which could be easily plugged into other recent works, e.g. vgg19, CSRNet \cite{CSRNet}, M-SFANet \cite{M-SFANet} and MAN \cite{MAN}. We compare the proposed methods with other state-of-the-art methods on six public crowd datasets.


Table~\ref{table:Comparisons} reports the experiment results. By incorporating the AL-PAPM into MAN \cite{MAN}, our ``MAN+AL-PAPM" achieves better counting performance in the large-scale datasets (UCF-QNRF, JHU-CROWD++, and NWPU) compared to recent excellent works such as chfL \cite{chfL}, P2PNet \cite{p2pnet}, and MAN \cite{MAN}. Specifically, on the largest-scale and most challenging crowd counting dataset NWPU \cite{NWPU}, our ``MAN+AL-PAPM" achieves the best performance with 4.31\% MAE and 2.41\% MSE improvement compared with the state-of-the-art approach, MAN \cite{MAN}. On the smaller dataset ShTech A and ShTech B, our proposed ``P2P-PAPM'' also gains the best performances. 
Moreover, compared to original methods, the proposed HD-PAPM, AL-PAPM, and P2P-PAPM achieve better performances on each dataset. This results demonstrates the effectiveness of the proposed PAPM methods. The reason may be that the proposed PAPM improve the model's robustness to annotation displacement, resulting in counting accuracy improvement.

\begin{figure*}[t]
\begin{center}
\includegraphics[width=1.0\linewidth]{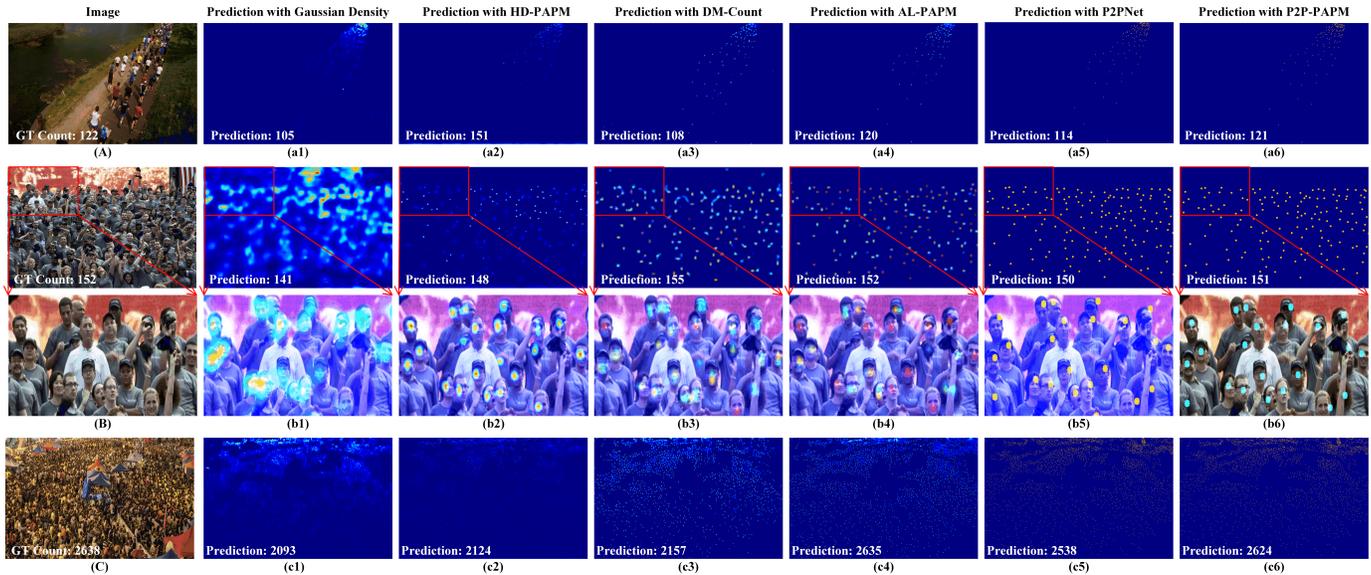}
\end{center}
   \caption{Comparison of the predictions of different methods. From left to right: input images, Gaussian density map method, HD-PAPM, DM-Count, AL-PAPM, P2PNet, and P2P-PAPM. We find that incorporating PAPM not only improves count accuracy, but also makes predicted responses and targets more consistent.}
\label{fig:vis}
\end{figure*}


\noindent \textbf{Visualization in crowd counting task.}
We first visualize the predictions of Gaussian density methods, DM-Count \cite{DM-Count}, and P2PNet \cite{p2pnet}. Then we compare the prediction results of the original methods and the methods combining PAPM. The detailed results have been presented in Figure \ref{fig:vis}.

In the column (B), we find that the response positions of Gaussian density methods, DM-Count, and P2PNet are random (e.g., face, eyes, or head). This randomness is attributed to annotation displacements introduced during the human annotation process. These offset annotations compel the network to learn features tied to the corresponding annotations, consequently impeding the acquisition of consistent features pertaining to the target.
Compared with the original methods, the response positions of the methods combining PAPM are more consistent, primarily aligning with the target's central region. This observation suggests that PAPM encourages the network to acquire consistent features related to the target, which in turn contributes to improvements in counting accuracy. In contrast to the original methods, our proposed approaches that incorporate PAPM yield count estimates that closely align with the ground-truth numbers. They also produce sharp PAPM which could localize the target well.

\begin{table}[htbp]
\caption{Localization performance on UCF-QNRF dataset.}
\begin{center}
\begin{tabular}{llll}
\hline & Precision & Recall & AUC \\
\hline MCNN \cite{MCNN} & $0.599$ & $0.635$ & $0.591$ \\
ResNet \cite{resnet} & $0.616$ & $0.669$ & $0.612$ \\
DenseNet \cite{densenet} & $0.702$ & $0.581$ & $0.637$ \\
Encoder-Decoder \cite{encoder} & $0.718$ & $0.630$ & $0.670$ \\
CL \cite{CL} & $0.758$ & $0.598$ & $0.714$ \\
BL \cite{BL} & ${0.767}$ & $0.654$ & ${0.720}$ \\
GL \cite{GL} & $\underline{0.782}$ & ${0.748}$ & $\underline{0.763}$ \\
\hline
Gaussian Densty & $0.605$ & ${0.670}$ & $0.623$ \\
HD-PAPM (ours) & $0.659$ & $0.731$ & $0.666$ \\
DM-Count \cite{DM-Count} & $0.731$ & $0.638$ & $0.692$ \\
AL-PAPM (ours) & $\textbf{0.797}$ & ${0.756}$ & $\textbf{0.781}$ \\
P2PNet \cite{p2pnet} & $0.712$ & $\underline{0.758}$ & $0.721$ \\
P2P-PAPM (ours) & ${0.744}$ & $\textbf{0.781}$ & ${0.745}$ \\
\hline
\end{tabular}
\label{localization}
\end{center}
\end{table}

\begin{table}[htbp]
\caption{Localization performance on NWPU-Crowd dataset.}
\begin{center}
\begin{tabular}{cccc}
\hline Method & Precision & Recall & F1-measure \\
\hline Faster RCNN \cite{Faster} & $\textbf{0.958}$ & $0.035$ & $0.068$ \\
TinyFace \cite{tiny} & $0.529$ & ${0.611}$ & $0.567$ \\
RAZNet \cite{RAZNet} & $0.666$ & $0.543$ & ${0.599}$ \\
D2CNet \cite{D2CNet} & $0.729$ & $0.662$ & ${0.700}$ \\
TopoCount \cite{topocount} & $0.695$ & ${0.687}$ & $0.691$ \\
CrossNet-HR \cite{CrossNet} & $0.748$ & $\textbf{0.757}$ & ${0.739}$ \\
GL \cite{GL} & ${0.800}$ & ${0.562}$ & ${0.660}$ \\
\hline
DM-Count & $0.738$ & $0.535$ & $0.618$ \\
AL-PAPM (ours) & $\underline{0.818}$ & ${0.602}$ & ${0.694}$ \\
P2PNet \cite{p2pnet} & $0.729$ & $0.695$ & $\underline{0.712}$ \\
P2P-PAPM (ours) & $0.769$ & $\underline{0.740}$ & $\textbf{0.756}$ \\
\hline
\end{tabular}
\label{localization_N}
\end{center}
\end{table}

\noindent \textbf{Localization.} As our PAPM methods produce sharp PAPMs, we followed \cite{GL} and evaluated the localization performance on the UCF-QNRF and NWPU datasets. 
For UCF-QNRF, the results presented in Table \ref{localization} show that our proposed AL-PAPM and P2P-PAPM outperform other methods, including the composition loss (CL) and P2PNet, which are specifically designed for localization. Additionally, our PAPM methods perform better than the original methods on each localization metric. The reason is that our PAPM assumes a substantial concentration of annotation points is centered within target regions. Thus, our proposed PAPM can be naturally used for localization. 
For localization results of NWPU in Table \ref{localization_N}, our ``P2P-PAPM'' achieves the best performance in F1-measure. And our ``AL-PAPM'' achieves the second best performance as quantified by Precision. Compared to DM-Count and P2PNet, our ``AL-PAPM'' and ``P2P-PAPM'' perform better on all the evaluation metrics. This indicates that the proposed PAPM benefits for localization. CrossNet-HR \cite{CrossNet} has the highest recall and F1-measure. The reason is that the CrossNet-HR is carefully designed for localization. While our method is designed for tolerating the annotation displacement.

\subsection{Vehicle Counting}
We also evaluate the performances of the HD-PAPM, AL-PAPM and P2P-PAPM on vehicle counting including TRANCOS \cite{V3}, CARPK \cite{CARPK} and PUCPR+ \cite{CARPK}.

\noindent \textbf{TRANCOS.}
For vehicle counting, we use the Grid Average Mean absolute Error (GAME) metric \cite{V3} on TRANCOS. With the GAME metric, we proceed to subdivide the image in $4^{L}$ non-overlapping regions, and compute the MAE in each of these sub-regions. The GAME is formulated as follows, 
\begin{equation}
G A M E(L)=\frac{1}{N} \sum_{i=1}^{N} \sum_{l=1}^{4^{L}}\left|\hat{g}_{i}^{l}-g_{i}^{l}\right|,
\end{equation}
where $N$ is the number of test images, $\hat{g}_{i}^{l}$ and $g_{i}^{l}$ are the estimated count and the ground truth in each sub-region, respectively. $L=\left\{ 0, 1, 2, 3 \right\}$ is a constant. The higher $L$, the more restrictive the GAME metric will be. Note that the MAE can be obtained as a particularization of the GAME when $L$ = 0.

The experiment results are shown in Table \ref{table7}. The proposed P2P-PAPM outperforms other state-of-the-art methods in MAE, GAME(1), GAME(2), and GAME(3) metrics. Compared with the original methods, the proposed methods after adding PAPM have achieved better performance in each evaluation metrics. The reason may be that our methods could tolerate the annotation displacement, resulting in counting accuracy improvement.

\begin{table}[ht]
\caption{Comparison of proposed methods with several state-of-the-art algorithms on TRANCOS dataset.}
\begin{center}
\begin{tabular}{c|c|c|c|c}
\hline Methods & MAE & GAME(1) & GAME(2) & GAME(3) \\
\hline Victor et al. \cite{Alpher52} & $13.76$ & $16.72$ & $20.72$ & $24.36$ \\
Onoro et al. \cite{Alpher50} & $10.99$ & $13.75$ & $16.09$ & $19.32$ \\
CSRNet \cite{SANet} & $3.56$ & $5.49$ & $8.57$ & $15.04$ \\
PSDDN \cite{PSDDN} & $4.79$ & $5.43$ & $6.68$ & $\underline{8.40}$ \\
KDMG \cite{JHU2} & $3.13$ & $4.79$ & $ {6.20}$ & $8.68$ \\
\hline 
Gaussian Density & $3.78$ & $6.97$ & $9.20$ & $17.87$ \\
HD-PAPM (ours) & $ {2.65}$ & $ {4.01}$ & $6.21$ & $9.69$ \\
DM-Count \cite{DM-Count} & $3.27$ & $5.07$ & $6.76$ & $10.96$ \\
AL-PAPM (ours) & $\underline{2.24}$ & $ \underline{3.51}$ & $\underline{5.18}$ & $ {9.03}$ \\
P2PNet \cite{p2pnet} & $3.06$ & $4.56$ & $6.32$ & $10.65$ \\
P2P-PAPM (ours) & $\textbf{2.02}$ & $\textbf{3.26}$ & $\textbf{4.86}$ & $\textbf{7.96}$ \\
\hline
\end{tabular}
\label{table7}
\end{center}
\end{table}

\noindent \textbf{PUCPR+ and CARPK.}
Regarding parked car counting on PUCPR+ and CARPK datasets in Table \ref{table8}, P2P-PAPM achieves the best performance in terms of MAE and MSE. Similarly, HD-PAPM and AL-PAPM outperform previous state-of-the-art approaches, except for KDMG on CARPK. It is worth noting that HD-PAPM and AL-PAPM relies on a simple VGG19 backbone, while KDMG employs an adaptive kernel-based density map generation framework, explaining its superior performance on CARPK. These results confirm the effectiveness of our proposed methods in accurately counting the number of vehicles in both parking lots and on roads.

\begin{table}[htbp]
\caption{Comparison of proposed methods with several state-of-the-art algorithms on PUCPR+ and CARPK dataset.}
\begin{center}
\begin{tabular}{c|c|c|c|c}
\hline \multirow{2}{*}{ Methods } & \multicolumn{2}{c|}{ PUCPR+ } & \multicolumn{2}{c}{ CARPK } \\
\cline { 2 - 5 } & MAE & MSE & MAE & MSE \\
\hline Faster R-CNN \cite{Alpher53} & $39.88$ & $47.67$ & $24.32$ & $37.62$ \\
YOLO \cite{Alpher54} & $156.00$ & $200.42$ & $48.89$ & $57.55$ \\
One-Look \cite{Alpher55} & $21.88$ & $36.73$ & $59.46$ & $66.84$ \\
LPN Counting \cite{CARPK} & $22.76$ & $34.46$ & $23.80$ & $36.79$ \\
YOLO9000 \cite{Alpher56} & $130.43$ & $172.46$ & $45.36$ & $52.20$ \\
RetinaNet \cite{Alpher57} & $24.58$ & $33.12$ & $16.62$ & $22.30$ \\
IEP Counting \cite{Alpher59} & $15.17$ & $-$ & $51.83$ & $-$ \\
Densely Packed \cite{Alpher60} & $7.16$ & $12.00$ & $6.77$ & $ {8.52}$ \\
ADMG \cite{JHU2} & $3.57$ & $5.02$ & $7.14$ & $8.59$ \\
KDMG \cite{JHU2} & $3.01$ & $4.38$ & $ {5.17}$ & $\underline{6.94}$ \\
\hline 
Gaussian Density & $ {5.67}$ & ${8.72}$ & ${8.95}$ & $13.67$ \\
HD-PAPM (ours) & $ {2.70}$ & ${3.70}$ & ${6.06}$ & $8.67$ \\
DM-Count & $ {3.46}$ & ${4.82}$ & ${6.36}$ & $7.42$ \\
AL-PAPM (ours) & $\underline{2.10}$ & $\underline{2.94}$ & $\underline{5.40}$ & ${7.34}$ \\
P2Pnet & $ {2.64}$ & ${3.40}$ & ${6.44}$ & $8.97$ \\
P2P-PAPM (ours) & $\textbf{2.02}$ & $\textbf{2.83}$ & $\textbf{5.22}$ & $\textbf{7.08}$ \\
\hline
\end{tabular}
\label{table8}
\end{center}
\end{table}

\begin{figure}[htbp]
\begin{center}
\includegraphics[width=1.0\linewidth]{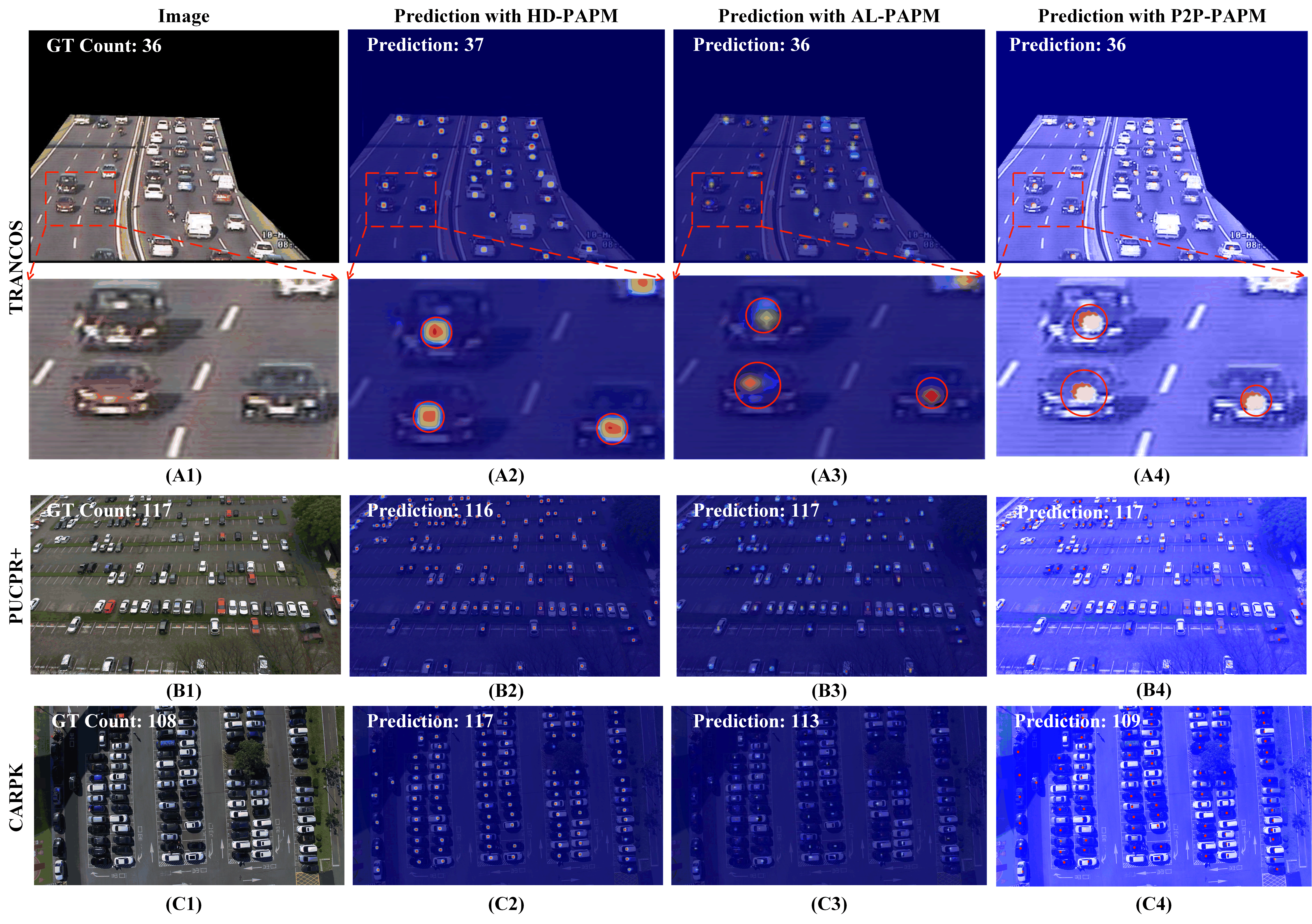}
\end{center}
   \caption{Visualization of estimated PAPMs and dilation maps on TRANCOS, PUCPR+, and CARPK.}
\label{fig:vis1}
\end{figure}

\noindent \textbf{Visualization of the estimated PAPM on vehicle counting task.}
In Figure \ref{fig:vis1}, we show the input images from TRANCONS \cite{V3}, PUCPR+ \cite{CARPK} and CARPK \cite{CARPK}, along
with the PAPMs predicted by our proposed methods. The first column is the input images. The second, third, and fourth columns are our predicted PAPMs by HD-PAPM, AL-PAPM and P2P-PAPM methods.
First, comparing the ground-truth and the predicted number, we could find that our HD-PAPM, AL-PAPM and P2P-PAPM, all have a good ability to accurately estimate the target number in different scenarios. This means that our learning target PAPM is general in different scenes. The reason may be that our PAPM concentrates on the annotation displacement that may be the same in labeling different targets. Second, as shown in (A2), (A3) and (A4), the response positions of HD-PAPM, AL-PAPM and P2P-PAPM are usually the center of the target. It means that our methods can produce responses consistent with the targets. One could be observed that the HD-PAPM has a higher response compared with AL-PAPM. The reason is that the AL-PAPM is an adaptively learning optimal transport framework. In contrast to HD-PAPM, the bandwidth of AL-PAPM is wider at 16, allowing it to transport probability mass of point annotation across a larger range. As a result, AL-PAPM is more robust to annotation noise, resulting in a more scattered and lower response visualization compared to HD-PAPM. However, despite the lower response, AL-PAPM achieves more accurate counting performance.

\subsection{General Object Counting}
To verify the generality of our proposed methods, we conducted experiments on the DOTA dataset, which contains different types of objects with varying shapes and sizes. We used the CSRNet \cite{CSRNet} as the backbone and trained it with different density maps including fixed kernel, adaptive kernel, DMG \cite{Adaptive-Density-Map2}, HD-PAPM, and AL-PAPM. We also evaluate P2PNet and P2P-PAPM on DOTA dataset. Specifically, we used 690 images with 6 classes with object numbers larger than 10, labeled as ``6 classes", as in \cite{Adaptive-Density-Map2} and 1869 high-resolution images with 18 classes, labeled as ``18 classes".
The results in Table \ref{table9} show that our proposed HD-PAPM outperforms fixed and adaptive Gaussian density methods, confirming that our proposed PAPM learning target is generalizable to most counting tasks. Even compared with the superior Kernel-based Density Map Generation (KDMG) method, our AL-PAPM achieves lower MAE and MSE. Combining stronger P2PNet, our proposed P2P-PAPM achieves the best performances in all settings. 

\begin{table}[ht]
\caption{Experiment results on DOTA dataset.}
\begin{center}
\begin{tabular}{c|c|c|c|c}
\hline \multirow{2}{*}{ Methods } & \multicolumn{2}{c|}{6 Classes} & \multicolumn{2}{c}{18 Classes} \\
\cline { 2 - 5 } & MAE & MSE & MAE & MSE \\
\hline 
\text {Gaussian Density}($\sigma=4$) & 4.82 & 10.17 & 18.35 & 58.36 \\
\text {Gaussian Density}($\sigma=16$) & 5.10 & 9.02 & 17.52 & 38.97\\
\text {Adaptive Gaussian Density} & 6.05 & 8.95 & 20.34 & 60.45\\
\text {ADMG \cite{Adaptive-Density-Map2}} & 4.42 & 8.38 & $-$ & $-$\\
\text {KDMG \cite{Adaptive-Density-Map2}} &  {3.65} &  {7.44} & $-$ & $-$\\
DM-Count \cite{DM-Count} & 4.23 & 7.86 & 15.71 & 34.56 \\
P2PNet \cite{p2pnet} & 4.02 & 8.44 & 14.24 & 36.76 \\
\hline
\text {HD-PAPM (ours)} & 4.36 & 8.09 &  {15.21} &{34.67}\\
\text {AL-PAPM (ours)} &\underline{3.05} &\underline{6.65} &\underline{13.81} &\underline{31.45}\\
\text {P2P-PAPM (ours)} & \textbf{2.85} & \textbf{6.11} & \textbf{10.42} & \textbf{28.76} \\
\hline
\end{tabular}
\label{table9}
\end{center}
\end{table}

\subsection{Ablation Study}
In this subsection, we conduct an ablation study to analyze the tunable parameters, choose a suitable transport cost function, and compare our proposed AL-PAPM with other loss functions. All experiments are conducted with vgg19 backbone.

\subsubsection{Parameter Analysis}
The proposed methods have several tunable parameters: bandwidth $\sigma$ and the shape parameter $s$ of GGD in the HD-PAPM; The bandwidth $\sigma$, the shape parameter $s$ of the GGD-L2 combination transport cost function, and the weights $\lambda$ in the proposed AL-PAPM; The bandwidth $\sigma$, the shape parameter $s$ of GGD-based cost matrix in P2P-PAPM.
In this section, we conduct a series of experiments to study the sensitivity issues of the parameters.

\noindent \textbf{Effect of the tunable parameters in HD-PAPM.}
To evaluate the effect of the tunable parameters, the bandwidth $\sigma$ and the parameter $s$ in the proposed HD-PAPM, we first fix bandwidth $\sigma$ to 4 and tune the parameter $s$ from  0.5, 1, 2, 4, 8 to 16, on ShTech A dataset. As shown in Table \ref{tab:s Setting}, $s = 8$ outperforms other weight values. Then we fix $s$ to 8 and tune bandwidth $\sigma$ from 2, 4, 8, 16 to 32. As shown in Table \ref{tab:bandwidth Setting1},  $\sigma = 4$ outperforms other bandwidth values. 
We found that a bandwidth parameter $\sigma = 4$ is a suitable displacement range for annotators to mark point annotations in the HD-PAPM method. We believe that labeling displacement is acceptable within this range. As shown in Figure \ref{fig:s}, we observed that the sharpness near the origin was not smooth enough when $s$ was set to $0.5, 1, 2,$ or $4$. This contradicts our assumption that people have an equal probability of marking points in the target area. Conversely, when $s = 16$, the slope of the curve is too high, suggesting that people's annotation range is fixed, which is contrary to reality. After conducting a thorough analysis, we found that the curve of $s = 8$ not only satisfies people's tendency to label on the target region but also tolerates labeling displacement. Thus, we set $\sigma = 4$ and $s = 8$ for HD-PAPM on all datasets.


\begin{table}[htbp]
\caption{Effect of different parameter $s$ setting and bandwidth $\sigma$ setting on ShanghaiTech Part A dataset.}
\begin{center}
$$
\begin{array}{ccc}
\hline \text {Parameter $s$} & \text { MAE } & \text { MSE } \\
\hline 
0.5 & 75.2 & 122.6 \\
1 & 68.4 & 113.2 \\
2 & 65.2 & 108.6 \\
4 & 65.7 & 104.2\\
8 & \textbf{62.3} & \textbf{101.2} \\
16 & 65.9 & 103.9 \\
\hline
\end{array}
$$
\end{center}
\label{tab:s Setting}
\end{table}

\begin{table}[htbp]
\caption{Effect of different bandwidth $\sigma$ setting on ShanghaiTech Part A dataset.}
\begin{center}
$$
\begin{array}{ccc}
\hline \text {  Bandwidth $\sigma$ Setting } & \text { MAE } & \text { MSE } \\
\hline 
2 & 64.2 & 103.8 \\
4 & \textbf{62.3} & \textbf{101.2} \\
8 & 68.5 & 105.6 \\
16 & 67.3 & 108.8 \\
32 & 71.1 & 111.3 \\
\hline
\end{array}
$$
\end{center}
\label{tab:bandwidth Setting1}
\end{table}

\begin{figure}[ht]
\begin{center}
\includegraphics[width=0.9\linewidth]{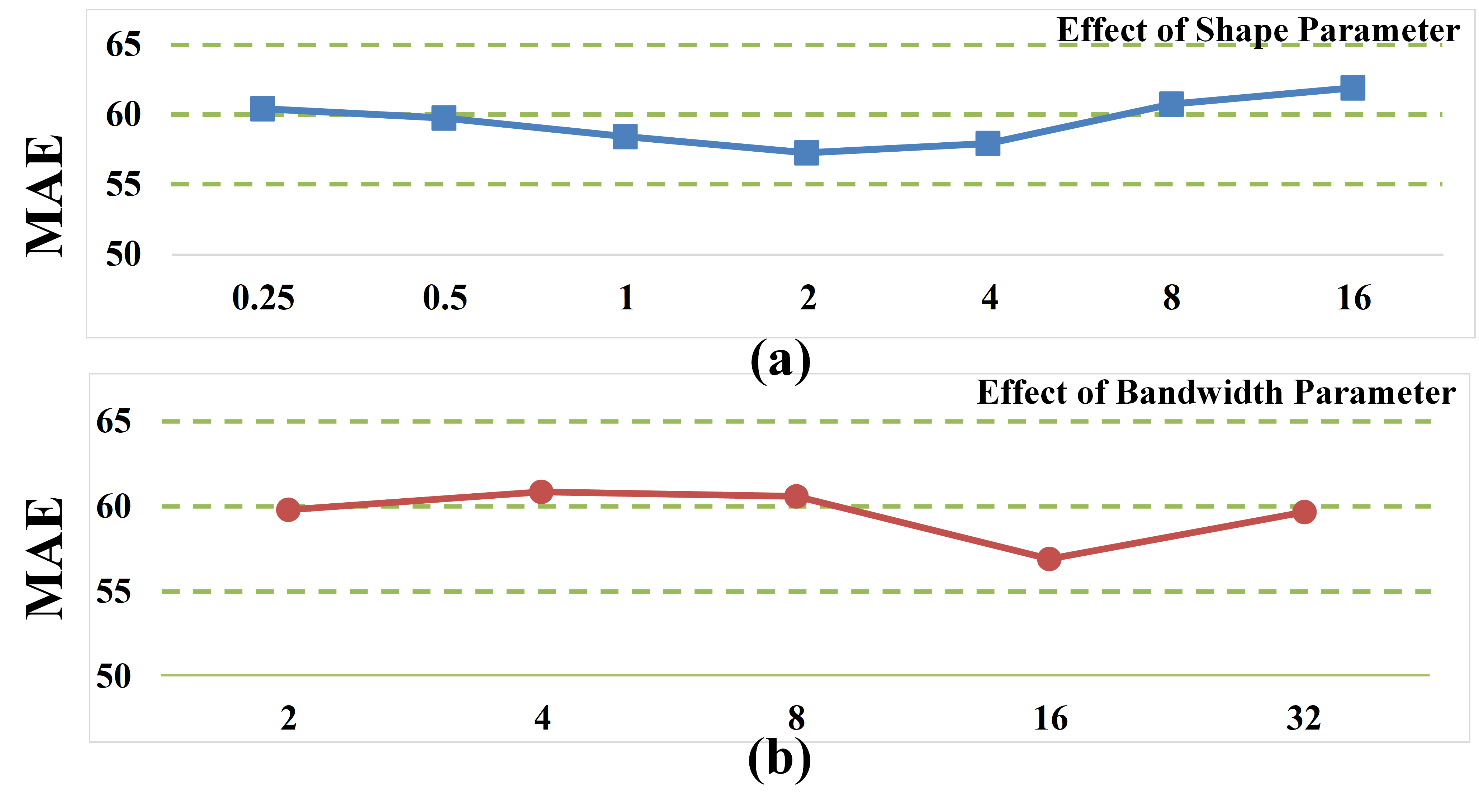}
\end{center}
   \caption{The curves of testing results for the bandwidth $\sigma$ and the shape parameter $s$ in AL-PAPM on ShanghaiTech part A.}
\label{fig:AL parameters}
\end{figure}

\noindent \textbf {Effect of the tunable parameters in AL-PAPM.}
To evaluate the effect of the tunable parameters, the bandwidth $\sigma$ and the shape parameter $s$ in the proposed AL-PAPM, we first fix bandwidth $\sigma$ to 4 and tune the parameter $s$ from  0.25, 0.5, 1, 2, 4, 8 to 16, on ShTech A dataset. As shown in Figure \ref{fig:AL parameters} (a), $s = 2$ outperforms other weight values in MAE. Then we fix $s$ to 2 and tune bandwidth $\sigma$ from 2, 4, 8, 16 to 32. As shown in Figure \ref{fig:AL parameters} (b),  $\sigma = 16$ outperforms other bandwidth values. In AL-PAPM, when $s=2$, the image with $\sigma=16$ is more consistent with our hypothesis: people will mark points within a certain range, and this displacement range should not be very large (about a distance of more than a dozen pixels). As shown in Figure \ref{fig:als}, When $\sigma=16$, the images of $s=0.5, 1$ could not reflect that people will mark points within a certain range. The reason is because their transmission cost is not 0 when the abscissa is 0 to 15. While for $s=4, 8$, transmission cost closes to 0 when the abscissa is 0 to 40, which is contrary to our assumption. $\sigma=16, s=2$ makes the function similar to the hypothesis, so it can obtain a better experimental result. Therefore, we choose $\sigma = 16$, $s = 2$ for AL-PAPM on all datasets.

To evaluate the effect of the weight set in the proposed AL-PAPM, we set different magnitude weights, from $0.01, 0.1, 1$ to 10, on ShTech A dataset. As shown in Table \ref{tab:Weight Setting}, $\lambda=0.1$ outperforms other weight values. Therefore, we set $\lambda=0.1$ for experiments on all datasets.

\begin{table}[htbp]
\caption{Effect of different $\lambda$ weight setting on ShanghaiTech Part A dataset.}
\begin{center}
$$
\begin{array}{ccc}
\hline \text {  Weight Setting ($\lambda$) } & \text { MAE } & \text { MSE } \\
\hline 0.01 & 59.7 & 95.5 \\
0.1 & \textbf{57.1} & \textbf{92.5} \\
1 & 62.7 & 97.8 \\
10 & 66.9 & 105.7 \\
\hline
\end{array}
$$
\end{center}
\label{tab:Weight Setting}
\end{table}

\noindent \textbf {Effect of the tunable parameters in P2P-PAPM.} The hyper parameters $\sigma$ and $s$ in P2P-PAPM are the same to those in AL-PAPM. As a result, we choose $\sigma = 16$, $s = 2$ for P2P-PAPM on all datasets. While for other hyper-parameters, we set them the same to P2PNet \cite{p2pnet} for a fair comparison.



\begin{table}[htbp]
\caption{Effect of different transport cost functions. We follow the experiment settings in \cite{DM-Count} and conduct these experiments on ShTech A and UCF-QNRF dataset with VGG19 backbone.}
\setlength\tabcolsep{3pt}
\begin{center}
\begin{tabular}{c|c|c|c|c}
\hline \multirow{2}{*}{ Component } & \multicolumn{2}{c|}{ShTech A} & \multicolumn{2}{c}{UCF-QNRF} \\
\cline { 2 - 5 } & MAE & MSE & MAE & MSE \\
\hline
Similarity Counting (SC) loss & $66.7$ & $105.9$ & $94.9$ & $167.4$ \\
SC+L2 transport cost \cite{DM-Count} & $59.7$ & $95.7$ & $85.6$ & $148.3$ \\
SC+PG transport cost \cite{GL} & $60.5$ & $94.6$ & $83.6$ & $146.0$ \\
\hline 
SC+GGD-L2 combination transport cost & $57.1$ & $92.5$ & $81.2$ & $141.9$ \\
\hline
\end{tabular}
\label{tab:Component}
\end{center}
\end{table}

\subsubsection{The effect of different transport cost functions}

In Table \ref{tab:Component}, we evaluate the effect of different transport cost functions on the ShTech A and UCF-QNRF datasets by comparing our method with other cost functions, including the squared Euclidean distance (L2) in \cite{DM-Count} and the Perspective-Guided (PG) Transport Cost in \cite{GL}. We observe that the transport cost functions have a significant impact on the counting performance. Our proposed GGD-L2 combination transport cost function achieves the best results. In comparison, the classic L2 transport cost function and PG transport cost function are less effective than our GGD-L2 combination function, which may be due to their failure to account for annotation displacement. As illustrated in Figure \ref{fig:kernel function} (c), our GGD-L2 combination transport cost function suggests that the transmission cost within a certain range is close to 0. Consequently, it could tolerate the displacement of the annotated locations in the target region, resulting in the best results.

As the log GGD, i.e., $c = (||p_i-a_j||\sigma)^s$ in Figure \ref{fig:kernel function} (b) can give a similar shape as the GGD-L2 combination function in Figure \ref{fig:kernel function} (c), an ablation study has been conducted to justify the design choice of our proposed cost function. The detail results in Table \ref{tab:s-Setting} show that GGD-L2 combination transport cost function achieves the best performance. The GGD-L2 combination transport cost function in Figure \ref{fig:kernel function} (c) has a transmission cost that is close to 0 within a specific range, as illustrated. This property allows it to accommodate the displacement of annotated locations in the target region. While the log GGD transport cost functions in Figure \ref{fig:kernel function} (b) have a weaker ability to tolerate displacement. Thus, the GGD-L2 combination transport cost function is used for training with OT loss.

\begin{table*}[htbp]
    \centering
    \caption{Robustness to annotation noise. These experiments are conducted on ShanghaiTech Part A dataset.}
    \begin{tabular}{l|c|c|c|c|c}
    \hline MAE/MSE & 0 & 4 & 8 & 16 & 32 \\
    \hline Gaussian Density & 68.6/110.1 & 72.2/116.7 & 78.1/127.4 & 80.4/129.3 & 92.3/144.6 \\
    \hline HD-PAPM (ours) & 62.3/101.2 & 63.1/104.9 & 64.8/109.2 & 65.5/114.5 & 74.4/127.8 \\
    \hline DM-Count \cite{DM-Count} & 59.7/95.7 & 61.9/98.2 & 63.6/103.5 & 65.8/107.5 & 70.4/115.3 \\
    \hline AL-PAPM (ours) & 57.1/92.5 & 57.8/93.8 & 58.9/95.6 & 60.1/97.7 & 61.8/99.0 \\
    \hline NoiseCC \cite{NoiseCC} & 61.9/99.6 & 63.1/101.8 & 64.8/104.8 & 65.5/106.9 & 67.9/112.6 \\
    \hline P2PNet \cite{p2pnet} & 52.7/85.1 & 53.8/87.9 & 56.7/93.4 & 60.1/98.8 & 68.8/107.4 \\
    \hline P2P-PAPM (ours) & \textbf{51.4}/\textbf{82.8} & \textbf{52.1}/\textbf{84.2} & \textbf{53.3}/\textbf{87.6} & \textbf{56.7}/\textbf{92.0} & \textbf{61.0}/\textbf{98.8} \\
    \hline
    \end{tabular} 
    \label{tab:noise}
\end{table*}

    \begin{table}[htbp]
\caption{Effect of different transport cost functions. The experiments are conducted in ShanghaiTech Part A dataset with VGG19 backbone.}
\begin{center}
$$
\begin{array}{ccc}
\hline \text {Transport Cost Function} & \text { MAE } & \text { MSE } \\
\hline 
c = (||p_i-a_j||/16) & 59.2 & 97.7 \\
c = (||p_i-a_j||/16)^2 & 57.5 & 97.5 \\
c = (||p_i-a_j||/4)^4 & 58.2 & 96.1 \\
c = (||p_i-a_j||/16)^4 & 57.1 & 93.8 \\
c = (||p_i-a_j||/64)^4 & 57.4 & 94.2 \\
c = (||p_i-a_j||/16)^6 & 59.6 & 94.4 \\
c = ||p_i-a_j||*exp(||p_i-a_j||^{2}/(2*16^{2})) & 57.6 & 94.4 \\
c = ||p_i-a_j||^{2}*exp(||p_i-a_j||^{2}/(2*16^{2})) & \textbf{57.1} & \textbf{92.5} \\
\hline
\end{array}
$$
\end{center}
\label{tab:s-Setting}
\end{table}

\subsubsection{Comparison with different loss functions}

In Table \ref{tab:different backbones}, we compare our proposed loss function with different loss functions (AL-PAPM) using different backbone networks. The pixel-wise L2 loss function measures the pixel difference between the predicted density map and the ``ground-truth" density map. The BL \cite{BL} uses a point-wise loss function between the ground-truth point annotations and the aggregated dot prediction generated from the predicted density map. The NoiseCC models \cite{NoiseCC} the annotation noise using a random variable with Gaussian distribution and derives a probability density Gaussian approximation as a loss function. DM-Count \cite{DM-Count} uses balanced OT with an L2 cost function, to match the shape of the two distributions. GL \cite{GL} is an unbalanced optimal transport (UOT) framework with a perspective-guided transport cost function.

Our proposed AL-PAPM can be easily incorporated into existing crowd counting models, such as DM-Count and GL. By adding the GGD-L2 combination transport cost function, we obtain two enhanced models, ``DM-Count+AL-PAPM" and ``GL+AL-PAPM". The experimental results, presented in Table \ref{tab:different backbones}, demonstrate that ``GL+AL-PAPM" achieves the best performance among all loss functions when combined with the GL architecture. Furthermore, our methods ``DM-Count+AL-PAPM" and ``GL+AL-PAPM" outperform the traditional L2 loss function since we directly use point annotations for supervision, rather than designing a hand-crafted intermediate representation as a learning target. Compared to other methods that use point annotations for supervision, such as BL, DM-Count, and GL, our proposed method ``GL+AL-PAPM" achieves superior performance across all network architectures, as it could tolerate the displacement of the annotated locations in the target region. 

\begin{table}[htbp]
\caption{Performances of loss functions using different backbones on UCF-QNRF dataset. Our proposed method outperforms other loss functions.}
\begin{center}
$$
\begin{array}{llll}
\hline \multirow{2}*{\text { Methods }} & {\text { VGG19  }}& {\text { CSRNet  }}& {\text { MCNN } } \\
& \text { MAE} / \text {MSE } & \text { MAE} / \text {MSE } & \text { MAE} / \text {MSE } \\
\hline \text {L2 } & 98.7  / 176.1 & 110.6  / 190.1 & 186.4  / 283.6 \\
\text {BL \cite{BL} } & 88.8  / 154.8 & 107.5  / 184.3 & 190.6  / 272.3 \\
\text {NoiseCC \cite{NoiseCC}} & 85.8  / 150.6 & 96.5  / 163.3 & 177.4  / 259.0 \\
\text {DM-Count \cite{DM-Count} } & 85.6  / 148.3 & 103.6  / 180.6 & 176.1  / 263.3 \\
\text {DM-Count+AL-PAPM} & {81.2}  / {141.9} & {95.6}  / {162.7} & {157.5}  / {243.3} \\
\text {GL \cite{GL} } & 84.3  / 147.5 & 92.2  / 165.7 & 142.8  / 227.9 \\
\text {GL+AL-PAPM} & \textbf{80.1}  / \textbf{140.2} & \textbf{90.6}  / \textbf{160.3} & \textbf{138.5}  / \textbf{219.4} \\
\hline
\end{array}
$$
\end{center}
\label{tab:different backbones}
\end{table}


\subsection{Robustness to annotation noise}
Since the proposed PAPM considers the annotation displacement, we experiment on ShanghaiTech A to verify its robustness to annotation noise. To be specific, we follow previous work \cite{NoiseCC} and generate a noisy dataset by moving the annotation points by $\{4, 8, 16, 32\}$ pixels. Then we train the vgg19 backbone with different learning targets including Gaussian density map with per-pixel loss \cite{CSRNet}, HD-PAPM with per-pixel loss, NoiseCC \cite{NoiseCC}, , DM-Count \cite{DM-Count}, and AL-PAPM. As depicted in Table \ref{tab:noise}, it's evident that the counting errors for these approaches increase as the level of annotation noise escalates. Notably, when comparing the proposed methods that incorporate PAPM with the original methods, it's apparent that the former achieve significantly lower MAE/MSE values when confronted with varying degrees of annotation displacement. This observation suggests that the proposed PAPM enhances the robustness of these methods to annotation noise. 

\section{Conclusion}
In this paper, we introduce a novel learning target called the Point Annotation Probability Map (PAPM) for object counting tasks. PAPM is based on the fundamental assumption that each annotation point within the target region contributes equally to the counting task. To achieve this, we employ a Generalized Gaussian Distribution (GGD) function with tunable bandwidth and shape parameters in PAPM. This allows PAPM to assume consistent annotation probabilities within the target region. This property of PAPM makes it robust to annotation displacement.
PAPM serves as a genera; concept that can be seamlessly integrated with various counting methodologies. We combine PAPM with Gaussian density, DM-Count, and P2PNet, resulting in HD-PAPM, AL-PAPM, and P2P-PAPM, respectively. These proposed methods show improved robustness to annotation displacement and subsequently lead to enhanced counting accuracy when compared to the original methods. Extensive experiments validate the effectiveness and superiority of the proposed PAPM-based approaches.
\ifCLASSOPTIONcompsoc
  \section*{Acknowledgments}
\else
  \section*{Acknowledgment}
\fi
This work was supported by the National Natural Science Foundation of China under Grant No. 62073257, and the Key Research and Development Program of Shaanxi Province of China under Grant No. 2022GY-076.

\ifCLASSOPTIONcaptionsoff
  \newpage
\fi



\bibliographystyle{IEEEtran}
\bibliography{egbib}
\end{document}